\PassOptionsToPackage{table}{xcolor}
\PassOptionsToPackage{pdfa}{hyperref}
\documentclass[acmsmall,authorversion,nonacm]{acmart}
\hypersetup{pdfapart=2, pdfaconformance=u}
\usepackage{amsfonts}
\usepackage{algorithmic}
\usepackage{algorithm}
\usepackage[caption=false,font=normalsize,labelfont=sf,textfont=sf]{subfig}
\usepackage{textcomp}
\usepackage{stfloats}
\usepackage{url}
\usepackage{verbatim}
\usepackage{graphicx}
\usepackage{balance}
\usepackage{soul}
\usepackage{multirow}
\usepackage[titletoc]{appendix}

\AtBeginDocument{%
  }

\setcopyright{acmlicensed}
\acmDOI{XXXXXXX.XXXXXXX}

\acmJournal{TIST}

\begin{document}
\title{Entropy Causal Graphs for Multivariate Time Series Anomaly Detection}

\author{Falih Gozi Febrinanto}
\email{f.febrinanto@federation.edu.au}
\affiliation{%
	\department{Institute of Innovation, Science and Sustainability}
  \institution{Federation University Australia, Australia and CSIRO's Data61}
  \city{Ballarat}
  \country{Australia}
}

\author{Kristen Moore}
\email{kristen.moore@data61.csiro.au}
\affiliation{%
  \institution{CSIRO's Data61}
  \city{Melbourne}
  \country{Australia}
}

\author{Chandra Thapa}
\email{chandra.thapa@data61.csiro.au}
\affiliation{%
  \institution{CSIRO's Data61}
  \city{Sydney}
  \country{Australia}
}

\author{Mujie Liu}
\email{mujie.liu@ieee.org}
\affiliation{%
	\department{Institute of Innovation, Science and Sustainability}
  \institution{Federation University Australia}
  \city{Ballarat}
  \country{Australia}
}

\author{Vidya Saikrishna}
\email{v.saikrishna@federation.edu.au}
\affiliation{%
	\department{Institute of Innovation, Science and Sustainability}
  \institution{Federation University Australia}
   \city{Ballarat}
  \country{Australia}
}

\author{Jiangang Ma}
\email{j.ma@federation.edu.au}
\affiliation{%
	\department{Institute of Innovation, Science and Sustainability}
  \institution{Federation University Australia}
  \city{Melbourne}
  \country{Australia}
}

\author{Feng Xia}
\authornote{Corresponding Author}
\email{f.xia@ieee.org}
\affiliation{%
	\department{School of Computing Technologies}
  \institution{RMIT University}
  \city{Melbourne}
  \country{Australia}
}

\renewcommand{\shortauthors}{Febrinanto et al.}


\begin{abstract}
Many multivariate time series anomaly detection frameworks have been proposed and widely applied. 
However, most of these frameworks do not consider intrinsic relationships between variables in multivariate time series data, thus ignoring the causal relationship among variables and degrading anomaly detection performance.  
This work proposes a novel framework called CGAD, an entropy Causal Graph for multivariate time series Anomaly Detection.
CGAD utilizes transfer entropy to construct graph structures that unveil the underlying causal relationships among time series data. 
Weighted graph convolutional networks combined with causal convolutions are employed to model both the causal graph structures and the temporal patterns within multivariate time series data. 
Furthermore, CGAD applies anomaly scoring, leveraging median absolute deviation-based normalization to improve the robustness of the anomaly identification process.
Extensive experiments demonstrate that CGAD outperforms state-of-the-art methods on real-world datasets with a 9\% average improvement in terms of three different multivariate time series anomaly detection metrics.

\end{abstract}




\maketitle

\section{Introduction}
\label{sec:introduction}

Advances in data collection and storage technologies have facilitated the widespread capture of multivariate time series data from sensors in complex systems. These systems span sectors such as manufacturing, energy, and transportation~\cite{liang2024foundation_timeseriessurvey}. Multivariate time series data analysis is a potent tool for discerning historical patterns and trends, and it plays a crucial role in a range of applications. Notably, it aids in the detection of atypical observations that might stem from system malfunctions, cyberattacks, or errors, thereby averting potential system failures~\cite{boniol2025vus_generalAD, xia2023deep}. This critical process, known as multivariate time series anomaly detection~\cite{jin20243survey_normallabelling}, holds significant importance in cybersecurity, where identifying and addressing these anomalies promptly is vital for maintaining system integrity and security. Its application is widespread, encompassing diverse fields such as critical infrastructure protection, surveillance, computer networks, finance, and healthcare, demonstrating its versatility and critical role in various domains~\cite{darban2022deep_timeseries_deeplearning, ho2025graph}.

In the field of multivariate time series anomaly detection, deep learning techniques have gained widespread adoption for their exceptional performance, often surpassing traditional statistical methods in terms of accuracy and effectiveness~\cite{boniol2025vus_generalAD, darban2022deep_timeseries_deeplearning}. Deep learning models in this context are typically trained using unsupervised techniques, operating under the assumption that the training dataset is devoid of anomalies and maintains a high level of cleanliness~\cite{han2023anomaly_normal, jin20243survey_normallabelling}. This assumption is used to construct an accurate model representation of normal patterns. During inference, the model is presented with new samples, encompassing both normal data and potential anomalies. Anomaly scores are then computed based on the deviation between the model's output and the established ground truth. These scores are used to determine whether the instances are anomalies or not~\cite{su2019robust_omnianomaly, tuli2022tranad}. However, the enhancement of anomaly detection performance is often impeded by a common limitation in deep learning techniques: they tend to process time series data in isolation without accounting for the inter-variable relationships inherent in multivariate datasets~\cite{deng2021graph_GDN}.

Recent studies in multivariate time series anomaly detection leverage graph learning techniques to model interdependencies among variables~\cite{jin20243survey_normallabelling}. This direction has gained significant attention and has yielded promising results to enhance the accuracy of anomaly detection~\cite{ren2023graph_graphanomaly, deng2021graph_GDN,  velivckovic2018graph_gat}.
Graph neural networks (GNNs) are powerful models in graph learning and are extensively utilized to tackle diverse downstream tasks in domains such as social networks and transportation systems~\cite{xia2025graph, xia2021graph, zhou2022graph_gnn}.
To effectively leverage GNNs for multivariate time series data, particularly in contexts like environmental monitoring, industrial automation, or smart cities, where sensors are used to collect various data points, it's essential to represent this data in a graph-structured format. In this structure, each sensor, functioning as a data collection point for specific variables like temperature, pressure, or movement, is represented as a node, and their relationships are depicted as edges~\cite{chen2024graph_graphtimeseries}. However, establishing an appropriate graph structure initially poses a significant challenge~\cite{febrinanto2023efficient}. Consequently, the role of automatic graph generation techniques becomes pivotal~\cite{li2024gslb_graphstructurelearning, shen2025towards}. These techniques are designed to create graph structures that intricately represent the complex interrelationships among variables in multivariate time series data, thereby facilitating more insightful analysis~\cite{jin20243survey_normallabelling}.

\textbf{Limitations of Existing Methods}.
Several studies have introduced methods to generate graph structures to address the challenge of hidden relationships in multivariate time series data. These approaches focus on learning optimal structures as an integral part of the model training process~\cite{li2024gslb_graphstructurelearning, shen2025towards}. The first effort in this direction involves a sampling method known as the Gumbel-softmax~\cite{jang2017categorical_gumbel} to generate discrete adjacency matrices using random categorical vectors. These matrices help determine the probability of establishing a direct connection between two nodes~\cite{shang2020discrete_gts,chen2021learning_gta}. However, the Gumbel-softmax does not fully utilize the actual domain knowledge available in multivariate time series data, which introduces sampling bias. It can also lead to overfitting on the training data and a reduced ability to generalize when constructing graph structures for new multivariate time series data samples~\cite{chen2022balanced}.

An alternative approach involves cosine similarity (or distance metrics) to measure the similarity between time series representations through an end-to-end learning process. These methods control the sparsity of graphs by implementing the top-$k$ function, which selects only the $k$ most similar relations at every node (where the user chooses the parameter $k$)~\cite{wu2020connecting_mtgnn,deng2021graph_GDN}. However, this technique is less flexible since it limits the number of relations to $k$, which may overlook important but less prominent relationships within the graph that require more than $k$ relationships. Furthermore, it also results in a graph where all nodes have the same degree~\cite {chen2022balanced}. The construction of a fully connected graph has also been employed to represent graph structures in multivariate time series data~\cite{zhao2020multivariate_mtadgat}. However, in some cases, this approach may be less flexible when unnecessary and redundant information is included in the graph structure.

A key limitation of previous methods is their failure to consider causal relationships~\cite{cheng2024cuts_causality} among variables in multivariate time series data. Understanding these causal connections is crucial, as they reveal whether changes in one sensor affect others, offering deeper and more informative insights that are essential to incorporate~\cite{jiao2020quality_causality}. The previous correlation or similarity calculations only examine bivariate covariance and do not provide insights into causality or its direction among time series~\cite{tank2021neural_causality}. Furthermore, the learnable graph structure created through end-to-end learning resembles a black-box approach~\cite{li2025can_blackbox}, as it doesn't offer insight into the internal process of the graph structure. Consequently, this technique is inadequate when searching for causal relationships among time series data~\cite{liu2023explainable_blackbox}.

\setcounter{footnote}{0}

\textbf{Our Work}. To address the above-mentioned limitations, we propose a novel method called CGAD\footnote{The code is available at \url{https://github.com/falihgoz/CGAD}}, an entropy Causal Graph for multivariate time series Anomaly Detection. Our objective is to pre-construct the meaningful graph structure based on the causal relationships between the time series, thereby enabling more informative anomaly detection. Furthermore, to enhance the flexibility of the generated graph, we aim to automate the construction of the graph structure without imposing limits on the number of generated edge relations. First, we introduce a framework that leverages the causal discovery of multivariate time series to construct causal graphs, where the causality factor assesses whether one variable causes changes in the others. For this, we employ transfer entropy~\cite{schreiber2000measuring_transferentropy}, a comprehensive information theory-based approach, to uncover causality among variables in multivariate time series. Transfer entropy measures the amount of information transferred from one variable to another over time, enabling the construction of a graph structure with edge weights that reflect the causal relationships between variables~\cite{jiao2020quality_causality}. Additionally, the causality maps generated by transfer entropy offer enhanced visual interpretability~\cite{lindner2019comparative_transfermorethangranger}, which is beneficial for anomaly diagnosis.

CGAD employs a forecasting-based anomaly detection strategy to characterize anomalous events by calculating the deviation between forecast predictions and real observations. The CGAD framework is divided into three main parts: 1) \textbf{causal graph generation} to generate graph structures using causal discovery techniques in multivariate time series; 2) \textbf{weighted GNN forecasting} for learning future patterns of multivariate time series based on weighted graph convolution networks (GCNs)~\cite{welling2016semi_gcn} and causal convolution~\cite{yu2015multi_causalconv} to model the causal graph structures and temporal pattern of multivariate time series data; 3) \textbf{median deviation scoring} to assess deviations between forecasting results and actual observations as anomaly scores for detecting anomalies within specific time frames. It utilizes the median absolute deviation (MAD)~\cite{jamshidi2022detecting_mad} to standardize the anomaly scores and enhance robustness for comparing and identifying anomalous events.
 
\textbf{Contributions} In summary, this paper presents three main contributions:
    
\begin{enumerate}
    \item We propose a novel framework for multivariate time series anomaly detection called CGAD. This framework effectively captures intrinsic causal relationships among sensors or variables, providing more informative knowledge to enhance anomaly detection performance.
  
    \item We propose causal graph utilization to construct graph adjacency matrices as inputs for weighted GNNs to model multivariate time series data. This involves sampling to estimate pairwise time series causality based on transfer entropy and assigning the graphs' edge features.

    \item We conduct an extensive experiment on real-world datasets, and the results showcase that CGAD outperforms state-of-the-art approaches with a 9\% average improvement based on three different multivariate time series anomaly detection metrics.
\end{enumerate}

The remainder of this paper is organized as follows: Section 2 provides an overview of related work for this study. Section 3 introduces our proposed method, CGAD, and explains each component. Section 4 presents extensive experiments for evaluating our model and includes a description of qualitative analysis using visualizations of the graph structure and anomaly diagnosis within the CGAD framework. Finally, in Section 5, we conclude this paper.

\section{Related Work}
\subsection{Multivariate Time Series Anomaly Detection}

General machine learning techniques can be applied to detect anomalies in multivariate time series data. For instance, K-Nearest Neighbors (KNN)~\cite{angiulli2002fast_KNN} uses the $k$ nearest neighbors to assign anomaly scores, while Isolation Forest (IF)~\cite{liu2008isolation_IF} leverages a binary search tree structure to isolate sample points. A clustering-based method employs an extended Fuzzy C-Means algorithm~\cite{li2021clustering_clusteringbased}, groups similar subsequences, and reconstructs the data with optimized cluster centers to assign anomaly scores. These general machine-learning approaches are simple and fast. However, the lack of efficient strategies to model temporal patterns for each variable in multivariate time series data hinders their ability to process complex time series. To improve the performance of anomaly detection, recent techniques implement detection methods based on deep learning, which can be classified into two categories: reconstruction-based and forecasting-based methods~\cite{tuli2022tranad, chen2021learning_gta}.

\paragraph{\textbf{Reconstruction-based Methods}} The primary method in multivariate anomaly detection involves a reconstruction-based technique. This approach focuses on reconstructing normal sequences to minimize reconstruction loss, with outputs mirroring the input lengths. Anomaly detection is achieved by utilizing the reconstruction error as an anomaly score, setting thresholds to ascertain the presence of anomalies. Higher anomaly scores and probabilities indicate sequences that are more challenging to reconstruct. Notable implementations of this technique include models like OmniAnomaly~\cite{su2019robust_omnianomaly}, MAD-GAN~\cite{li2019mad_madgan}, USAD~\cite{audibert2020usad}, TranAD~\cite{tuli2022tranad}, and MTAD-GAT~\cite{zhao2020multivariate_mtadgat}.

MTAD-GAT employs graph learning techniques~\cite{xia2025graph, xia2021graph, zhou2022graph_gnn}, particularly graph attention networks (GAT)~\cite{velivckovic2018graph_gat}, to explicitly explore the inter-variable relationships in multivariate time series data. This approach uses GAT to discern correlations across the time series, represented within a fully connected graph that illustrates connections between all series components.

\paragraph{\textbf{Forecasting-based Methods}}
The forecasting-based method involves learning normal data sequences to predict their future values. During inference, sequences with potential anomalies are examined. Anomalies are identified based on the discrepancy between the forecasted data and the actual (ground truth) sequences. Significant deviations or incorrect predictions by the model indicate anomalies at specific time points. Some works that use the forecasting-based method for detecting anomalies include GDN~\cite{deng2021graph_GDN} and GTA~\cite{chen2021learning_gta}. Both methods use Graph Neural Networks (GNNs) to enhance the temporal analysis to predict future time steps. 

In this work, we specifically focus on the forecasting-based approach to maximize the utilization of causal relationships in multivariate time series data. Unlike the reconstruction-based approach, which reconstructs historical data to detect anomalies, the forecasting-based approach specializes in predicting the next timestamp, aiming to demonstrate cause-and-effect relationships between variables. We also employ a graph learning technique to model the spatial information of interdependencies between variables, leveraging the causal graph structures.

\subsection{Graph Generation Techniques}
To address these challenges and effectively capture the interdependencies between variables, recent works have introduced graph generation techniques for constructing an adjacency matrix based on initially unknown structures~\cite{li2024gslb_graphstructurelearning}. For example, GTS~\cite{shang2020discrete_gts} and GTA~\cite{chen2021learning_gta} employ distinct approaches, both leveraging the Gumbel-softmax~\cite{jang2017categorical_gumbel} to sample edge probabilities. This process results in the creation of a discrete adjacency matrix through an end-to-end learning procedure. However, it is worth noting that the Gumbel-softmax does not effectively incorporate domain knowledge when applied to multivariate time series data~\cite{chen2022balanced}. Additionally, the end-to-end learning process for generating graph structures may potentially lead to overfitting and reduced generalization when applied to new data.

GDN~\cite{deng2021graph_GDN} and MTGNN~\cite{wu2020connecting_mtgnn} develop learnable vector embeddings $h_i$ for each time series. Subsequently, the similarity between pairwise learnable vector embeddings $h_i$ and $h_j$ for all variables in the multivariate time series data is computed, often using a dot product. The top-$k$ highest similarity values are selected as relations in the graph to ensure sparsity in the learned graph representation. However, this constraint leads to a graph where all nodes have the same degree. It may overlook potential connections beyond the top-$k$, potentially resulting in a loss of contextual information and a limited representation of the node's neighborhood. Another approach, MTAD-GAT~\cite{zhao2020multivariate_mtadgat}, utilizes a fully connected graph to represent connections between all time series. However, this approach is also not flexible and may contain useless or redundant information. On the other hand, CauGNN~\cite{duan2022multivariate_caugnn} employs a transfer entropy graph based on time series causal analysis to construct the adjacency matrix before conducting graph representation learning.

Most current graph generation methods do not consider the causality factor between variables. CauGNN uses transfer entropy to build the graph structure to perform multivariate time series forecasting. However, it employs the dominant direction of information flow to calculate the difference between pairwise causal discovery calculations involving the subtraction of opposite directions, resulting in only one possible direction between pairwise variables. This hinders the ability to enhance the flexibility of relations when two variables influence each other. Our work has a different utilization for constructing relationships to enable flexibility and maintain causal information. Additionally, we provide a sampling approach to estimate causal discovery calculations, minimizing computing costs when dealing with large sequences of time series data.

\subsection{Causal Discovery for Time Series}

The graph generation strategy in CGAD is based on causal discovery, which considers the causal influence between variables in multivariate time series data as relations. Granger causality~\cite{granger1969investigating_gragercausality} is a popular method for measuring causal relationships between time series data. The Granger causality approach uses an auto-regression approach to test the statistical hypothesis of whether a particular time series can help predict the future effects of another time series. Another way to perform causal discovery is using transfer entropy~\cite{schreiber2000measuring_transferentropy}, an information-theoretic approach considered as generalized Granger causality. The main idea behind this technique is that a variable $x$ is causing $y$ when $y$ can be better sequentially compressed using a combination of past information in $x$ and $y$ rather than $y$ alone. In contrast to the Granger causality method, the transfer entropy approach is adept at capturing both linear and non-linear relationships~\cite{jiao2020quality_causality}. Despite high computational costs associated with transfer entropy, especially in cases involving a large sequence of time series data and a substantial number of variables~\cite{jiao2020quality_causality}, several methods have been proposed to accelerate and estimate transfer entropy calculations, including PyIF~\cite{ikegwu2020pyif}, RTransferEntropy~\cite{behrendt2019rtransferentropy}, and IDTxl~\cite{wollstadt2019idtxl}.

\begin{table}[!ht]
\tiny
\footnotesize	
  \centering
  \caption{Notations used in this paper.}
 \resizebox{\textwidth}{!}{\begin{tabular}{l|l|l|l}
\hline
Notations & Descriptions & Notations & Descriptions \\ \hline
$X$ & Multivariate time series data. & $\hat{D}$ & Diagonal degree matrix. \\
$\textbf{\textit{x}}_{:,t}$ & Values of multivariate time series at a time step $t$. & $\theta^{(l)}$ & Parameter matrix of corresponding layer $l$. \\
$\textbf{\textit{x}}_{i,:}$ & Univariate time series sequence in a node $i$. & $f(X, A)$ & GNN function with input $X$ and $A$. \\
$X^{\mathrm{train}}$ & Multivariate time series as a training sample. & $f(X, A, \Theta)$ & GNN forecasting function with input $X$, $A$, and $\Theta$. \\
$X^{\mathrm{test}}$ & Multivariate time series as a testing sample. & $\textbf{\textit{z}}$ & Causal convolution embedding vector. \\
$w, q, o$ & Window size for time series. & $\star$ & Convolution operator. \\
$I$ & Variable based on histogram encoding of time series $\textbf{\textit{x}}_{i,:}$. & $f_{1 \times k}$ & Convolution filter with dilation size $k$. \\
$i$ & A possible value in variable $I$. & $k$ & Kernel size of convolution. \\
$J$ & Variable based on histogram encoding of time series $\textbf{\textit{x}}_{j,:}$. & $m$ & A number of convolution layers. \\
$j$ & A possible value in variable $J$. & $r$ & A number of receptive fields. \\
$p(\cdot)$ & Probability density function. & $\textbf{\textit{h}}$ & A new vector embedding. \\
$H(I)$ & Information entropy function of variable $I$. & $tanh(\cdot)$ & A tangent hyperbolic activation function. \\
$H(I,J)$ & Joint entropy function of variable $I$ and $J$. & $\sigma(\cdot)$ & A sigmoid activation function. \\
$H(I|J)$ & Conditional entropy function of variable $I$ given $J$. & $\textup{loss}_{\mathrm{MSE}}$ & MSE loss function. \\
$TE_{J\rightarrow I}$ & Transfer entropy function from $J$ to $I$. & $\textup{error}_{i,t}$ & Deviation between actual value and prediction for node $i$ at time $t$. \\
$A$ & Graph adjacency matrix. & $\textup{med}_i$ & Median value of time series for node $i$. \\
$c$ & A constant to control weak relations. & $\textup{MAD}_i$ & Median absolute deviation (MAD) of time series for node $i$. \\
$\mathbf{\theta}$ & Weight parameter vector to be trained. & $\textbf{\textit{a}}_{i,t}$ & Anomaly score for node $i$ at time $t$. \\
$W,\Theta$ & Weight parameter matrix to be trained. & $s_{t}$ & Colective anomaly score overall nodes at time $t$. \\
$b$ & A scalar parameter to be trained. & $\textup{MAX}(\cdot)$ & Max aggregation function for the collective anomaly score. \\
$H^{(l+1)}$ & A new matrix embedding of layer $l+1$. & $y_{t}$ & Anomaly prediction at time $t$. \\
$\hat{A}$ & Normalized adjacency matrix. &  &  \\ \hline
\end{tabular}}
\label{tab::notation}
\end{table}

\section{Design of CGAD}
\label{sec:fram}
\subsection{Problem Formulation}
In this work, multivariate time series are denoted by $X\in \mathbb{R}^{N\times T}$, where $N$ represents the number of variables (\emph{e.g.}, number of sensors) and $T$ represents the length of each series. The values of multivariate time series at a specific time step $t$ are denoted as $\textbf{\textit{x}}_{:,t} \in \mathbb{R}^N$. In another view, the univariate time series sequence corresponding to the variable $i$ is denoted as $\textbf{\textit{x}}_{i,:} \in \mathbb{R}^T$. The system is classified as an anomaly at the time step $t$ if the cumulative anomaly score for all $N$ nodes in $\mathbf{x}_{:,t}$, denoted as $s_{t}$, exceeds the threshold. Eventually, the prediction based on the anomaly score $s_{t}$ is denoted as $y_{t} \in \{0,1\}$, where $0$ represents normal, and $1$ represents an anomaly.

\paragraph{\textbf{Anomaly Detection}} Following unsupervised anomaly detection, we assume that the multivariate time series data in training samples denoted $X^{\mathrm{train}}$, consist entirely of normal data, while the testing samples, denoted $X^{\mathrm{test}}$, contain some anomalies. A forecasting-based approach better represents the anomaly-free training dataset $X^{\mathrm{train}}$. The training process develops a model that predicts the next time step value $\textbf{\textit{x}}_{:,t}$ in the training samples, based on historical data $X_{t}^{\mathrm{train}}=\{\textbf{\textit{x}}_{:, t-w}, \dotsc, \textbf{\textit{x}}_{:,t - 1}\}$ for a given window size $w$. Then, during the inference phase, the trained model predicts the value $\textbf{\textit{x}}_{:,t}$ in the testing samples based on the historical data $X_{t}^{\mathrm{test}}=\{\textbf{\textit{x}}_{:,t-w}, \dotsc,\textbf{\textit{x}}_{:,t-1}\}$. The forecasting error, which is based on the difference between the forecasting result and the ground truth label, is then used to determine a cumulative anomaly score, denoted as $s_{t}$, for all $N$ nodes in  $\textbf{\textit{x}}_{:,t}$, to decide whether it is an anomaly or not. We summarize the notations in this work in Table~\ref{tab::notation}.

\begin{figure*}[!ht]
  \begin{center}
    \includegraphics[width=0.97\textwidth]{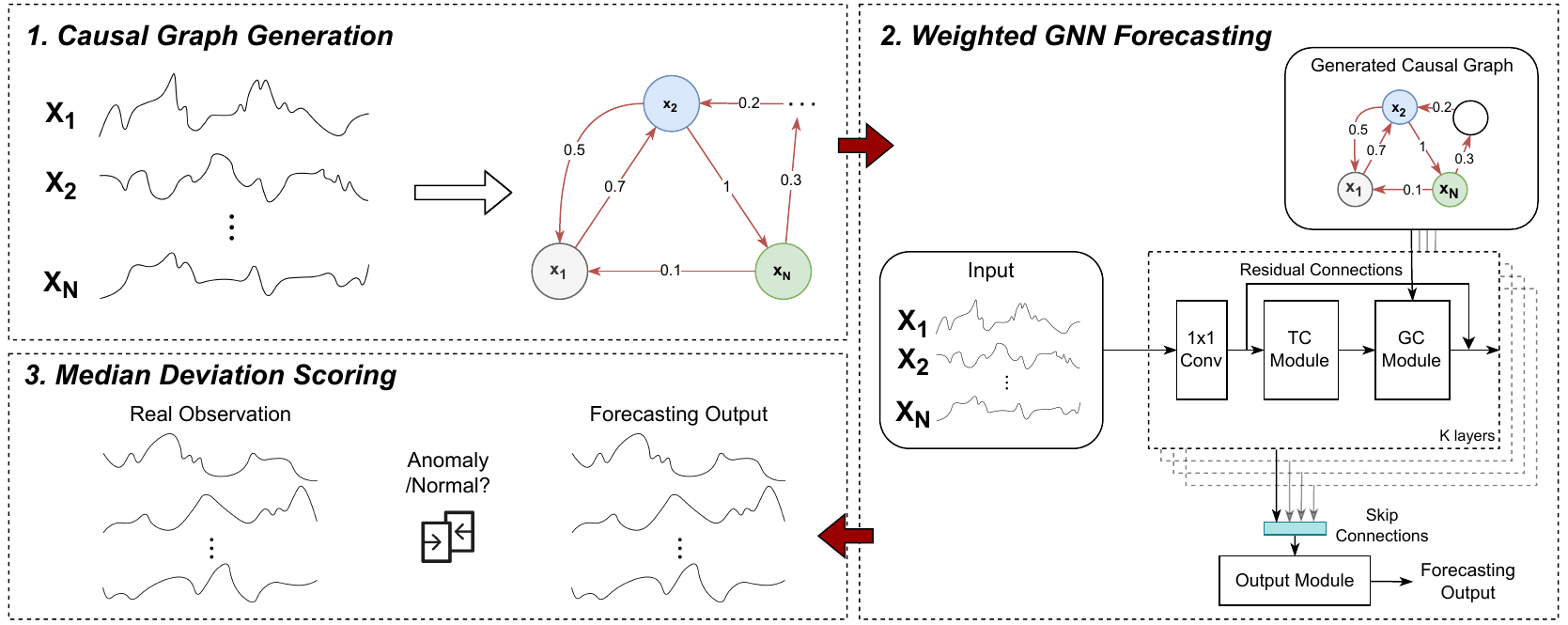}
  \end{center}
  \caption{An illustration of the proposed framework. The CGAD process begins with \textit{causal graph generation} based on transfer entropy~\cite{schreiber2000measuring_transferentropy} to establish causal relationships in multivariate time series data. \textit{Weighted GNN forecasting} is then performed for single-step forecasting, followed by \textit{median deviation scoring}, which computes the deviation between forecasting results and ground truths.}
  \label{img:fram}
  \vspace{-2pt}
\end{figure*}

\subsection{Framework Overview}
CGAD utilizes causality in multivariate time series data to form a graph. Our overall framework of CGAD is shown in Fig.~\ref{img:fram}. Based on our proposed method, there are three modules designed to perform anomaly detection:

\begin{enumerate}
  \item \textbf{Causal Graph Generation}. In the causal graph generation, we utilize transfer entropy~\cite{schreiber2000measuring_transferentropy} for discovering causality in multivariate time series data. Based on the causal discovery calculation, we develop an adjacency matrix that represents graph relations. We use the resulting adjacency matrix as input to the graph learning module in the CGAD model.
  \item \textbf{Weighted GNN Forecasting}. We adopt temporal analysis and capture temporal trends in each time series. Moreover, a GNN method is used to model the relationship between time series based on the causal graph adjacency matrix. This part aims to perform single-step forecasting for the multivariate time series.
  \item \textbf{Median Deviation Scoring}. This part is meant to identify anomalies in multivariate time series data by calculating anomaly scores based on forecasting errors. The intuition behind anomaly scoring is that if the forecasting result deviates from the ground truth, it will be labeled as an anomaly. We propose a standardization mechanism for anomaly scoring based on the Median Absolute Deviation (MAD). This mechanism transforms the scores into a common scale, enhancing the robustness for identifying anomalous events.
\end{enumerate}

\subsection{Causal Graph Generation}

This section introduces causal graph generation using transfer entropy (TE)~\cite{schreiber2000measuring_transferentropy}. It quantifies the level of uncertainty or randomness within a system and relies on the fundamental principle of Shannon entropy. We employ TE to calculate the causality factor between pairwise time series and to generate weighted causal graphs. TE is proven to be more precise and has visually more interpretable causality maps compared to other causal discovery methods, such as Granger causality~\cite{lindner2019comparative_transfermorethangranger}. 

There is a first data pre-processing stage in common modeling multivariate time series data, where data is normalized, typically using min-max normalization. Then, to compute Transfer Entropy (TE) in this work, the histogram approach is used to estimate the probability density function for each time series from its training data. To do this, the range of (uni-variate) time series $\textbf{\textit{x}}_{i,:}$ is divided into $D$ bins, and then the normalized counts for each bin estimate the probability that a time series measurement will fall into that bin. The information entropy, $H(I)$, can then be calculated as:
\begin{equation}
  \label{eq:te1}
  H(I) = -\sum_{i\in I} p(i)\log_2 p(i),
\end{equation}
where $p(\cdot)$ denotes the probability density function for the variable $I\in  \{1,\dotsc, D\}$ measuring which of the $D$ histogram bins the time series value $\textit{x}_{i,t}$ fell into. Equation~(\ref{eq:te1}) only calculates the information content of a single variable, whereas joint entropy measures the uncertainty associated with multiple variables. Consider another variable $J\in  \{1,..., D\}$ corresponding to the time series $\textbf{\textit{x}}_{j,:}$. The joint entropy, $H(I, J)$, captures the total uncertainty in the system by considering the interaction between $I$ and $J$:

\begin{equation}
  \label{eq:te2}
  H(I,J) := -\sum_{i \in I, j \in J} p(i,j)\log_2 p(i,j).
\end{equation} 
The joint entropy $H(I, J)$ quantifies the average amount of information required to specify the two variables' values precisely. The conditional entropy, $H(I|J)$, is related to the joint entropy and represents the information amount of $I$ under the condition when variable $J$ is known:
\begin{equation}
  \label{eq:te3}
  H(I|J) := -\sum_{i \in I, j \in J} p(i,j)\log_2 p(i|j).
\end{equation}
TE, denoted $TE_{J\rightarrow I}$, between variables $I$ and $J$ is measured by the information flow from $J$ to $I$~\cite{schreiber2000measuring_transferentropy}:
\begin{equation}
  \label{eq:te4}
  TE_{J\rightarrow I} := H(I_t|I_{t-1}) - H(I_t|I_{t-1},J_{t-1}),
\end{equation}
where the indices $t$ and $t-1$ represent the times $t$ and $t-1$, respectively. The full equation for TE can be expressed as:
\begin{equation}
  \label{eq:te5}
  TE_{J\rightarrow I} := \sum p(i_{t}, \textbf{\textit{i}}_{t-1}^{(q)}, \textbf{\textit{j}}_{t-1}^{(o)})\log_2\frac{p(i_t|\textbf{\textit{i}}_{t-1}^{(q)},\textbf{\textit{j}}_{t-1}^{(o)})}{p(i_t|\textbf{\textit{i}}_{t-1}^{(q)})},
\end{equation}
where $\textbf{\textit{i}}_{t-1}^{(q)}$ represents a possible value in the histogram encoding based on the time series $\textbf{\textit{x}}_{i,:}$, with the latest value at time $t-1$ and a window size of $q$. It covers the values of the time series as follows:  $[x_{i;t-1}, x_{i;t-2}, \ldots, x_{i;t-q+1}]$. Similarly, $\textbf{\textit{j}}_{t-1}^{(o)}$ comprises a possible value in the histogram encoding based on the time series $\textbf{\textit{x}}_{j,:}$, with the latest value at time $t-1$ and a window size of $o$. Thus, it includes the values of the time series as follows: $[x_{j;t-1}, x_{j;t-2}, \ldots, x_{j;t-o+1}]$. The most natural choice for the window size $q$ is $o$ or $1$~\cite{schreiber2000measuring_transferentropy}. We say that $J$ causes $I$ when $TE_{J\rightarrow I}$ is greater than 0. Higher values of $TE_{J\rightarrow I}$ means that time series $\textbf{\textit{x}}_{j,:}$ more strongly influences time series $\textbf{\textit{x}}_{i,:}$. 

\paragraph{\textbf{Causal Graph Utilization in CGAD}}
Here, we detail our transfer entropy-based graph generation approach and how it differs from CauGNN~\cite{duan2022multivariate_caugnn}. CauGNN uses a dominant direction of information flow to calculate the difference between both directions in pairwise TE calculation. Specifically, the calculation involves subtracting $TE_{J\rightarrow I}$ from $TE_{I\rightarrow J}$. However, this approach causes a loss of causal information if both directions have high causality values. We aim to preserve TE's natural value and maintain the influence direction between the two variables. Thus, it does not remove causal information in every direction. 

Once we get the causal relationship value based on the TE calculation, we can construct the graph's adjacency matrix $A \in \mathbb{R}^{N\times N}$, where $N$ is the number of nodes. Thus, the directed and weighted adjacency matrix is developed as follows:
\begin{equation}
  \label{eq:adj}
  A_{ij} =
  \begin{cases}
    TE_{\textbf{\textit{x}}_{j,:}\rightarrow \textbf{\textit{x}}_{i,:}} & \text{if $TE_{\textbf{\textit{x}}_{j,:}\rightarrow \textbf{\textit{x}}_{i,:}} > c$,} \\
    0           & \text{otherwise,}
  \end{cases}
\end{equation}
where $c$ is a control constant to prevent the development of unnecessary relations based on weak causality. The value of $c$ can be selected by the user based on the sparsity preference. However, the ideal number should be close to 0. 

To speed up the process of calculating TE, in this work, we implement a TE estimator based on Kraskov's method~\cite{kraskov2004estimating} that uses the k-nearest neighbors' strategy. We employ the PyIF~\cite{ikegwu2020pyif} library, an open-source Python library, to estimate TE with Kraskov's method, which has proven to be faster than other TE implementations. Additionally, we implement a sampling process since multivariate time series data have varying series lengths with potentially large time steps. This process involves randomly dividing the series into smaller chunks and calculating the average weighted adjacency matrix across all samples. This approach helps to perform causal discovery effectively with manageable series lengths. We summarize this process of causal graph generation in Algorithm~\ref{alg:1}.

\begin{algorithm}[tb]
\small
  \centering
  \caption{Causal Graph Generation Process}
  \label{alg:1}
  \begin{algorithmic}[1]
    \REQUIRE Multivariate time series data $X \in \mathbb{R}^{N \times T}$, window size $w$, number of samples $G$
    \ENSURE Adjacency matrix $A \in \mathbb{R}^{N \times N}$
    \STATE $X \leftarrow \text{minMaxScaler}(X)$ \COMMENT{Normalize $X$ using min-max scaler}
    \STATE $A \leftarrow \text{zeros}(G, N, N)$ \COMMENT{Initialize an empty adjacency matrix $A$ with dimensions $G \times N \times N$}
    \FOR{$g=1$ \TO $G$}
      \STATE $rand \leftarrow \text{random value from } (0, T - w - 1)$
      \FOR{$i=1$ \TO $N$}
        \FOR{$j=1$ \TO $N$}
          \IF{$i \neq j$}
            \STATE $A_{g,i,j} \leftarrow TE_{\textbf{\textit{x}}_{j,rand : rand + w} \rightarrow \textbf{\textit{x}}_{i,rand : rand + w}}$ \COMMENT{Calculate TE from node $j$ to node $i$ (Equation~\ref{eq:te5})}
          \ENDIF
        \ENDFOR
      \ENDFOR
    \ENDFOR
    \STATE $A \leftarrow \frac{1}{G} \sum_{g=1}^{G} A_g$ \COMMENT{Compute the adjacency matrix $A$ by averaging over the $G$ samples}
    \STATE Remove weak causal relationships in $A$ \COMMENT{Refer to Equation~\ref{eq:adj}}
  \end{algorithmic}
\end{algorithm}


\subsection{Weighted GNN Forecasting}
This section presents our proposed weighted graph neural network (GNN) forecasting approach to model spatial and temporal patterns in multivariate time series data. Our framework is based on the MTGNN model~\cite{wu2020connecting_mtgnn}, with modifications. Details of these modifications will be provided in later sections. 
Our weighted GNN forecasting consists of three main components: a weighted graph convolution module, a temporal convolution module, and a skip connection and output module. In addition, there is a residual connection from the input of the weighted graph convolution module added to the output of a graph convolution module to prevent the gradient vanishing problem.

\paragraph{\textbf{Weighted Graph Convolution Module}}
Our graph convolution module performs spatial analysis based on the generated causal graph. It is essential to extract node features and mine interdependencies between time series data. We define the graph convolution network (GCN) layer as in ~\cite{welling2016semi_gcn}:
\begin{equation}
  \label{eq:gcn}
  H^{(l+1)} = \sigma (\hat{D}^{-\frac{1}{2}}\hat{A}\hat{D}^{-\frac{1}{2}}H^{(l)}\Theta^{(l)}),
\end{equation}
where $H^{(l+1)}$ and $H^{(l)}$ denote the new embedding and the previous layer embedding before applying the graph convolutional operation, respectively. $\hat{A}=A + I$ is a normalized adjacency matrix with self-loops added to the original graph, $\hat{D}$ is the diagonal degree matrix of $\hat{A}$, $\Theta^{(l)}$ contains parameters of corresponding layer $l$, and $\sigma(\cdot)$ denotes the sigmoid function. The generated causal graphs outlined in the previous step are input into the GCN. Thus, we remove the graph structure learning module from MTGNN~\cite{wu2020connecting_mtgnn} to accommodate the causal graphs into the framework. Another modification is to employ the edge weight information as inputs of the graph convolution operator so that the adjacency matrix includes values other than 1 to represent the edge features.

We use a 2-layer GCN model~\cite{welling2016semi_gcn} as a good balance between ensuring the simplicity of the GCN while retaining the model's expressive power of the spatial dependencies in multivariate time series data. Our GCN model with input multivariate time series $X$ and a causal graph $A$ is expressed as:
\begin{equation}
  \label{eq:gcn_layer}
  f(X, A) = \textup{ReLU}(\widetilde{A}\,\textup{ReLU}(\widetilde{A}XW_0)W_1),
\end{equation}
where $\widetilde{A} = \hat{D}^{-\frac{1}{2}}\hat{A}\hat{D}^{-\frac{1}{2}}$ denotes the normalized adjacency matrix, $W_0 \in \mathbb{R}^{C\times H}$ represents the weight matrix from input to the first hidden layer, $W_1 \in \mathbb{R}^{H\times F}$ represents the weight matrix from the hidden layer to the output, $f(X, A) \in \mathbb{R}^{N\times F}$ represents the output with sequence length $F$, and $\textup{ReLU}(\cdot)$ is rectified linear unit, which is a standard activation function in deep neural networks.
\begin{figure}[h]
  \begin{center}
    \includegraphics[width=0.6\textwidth]{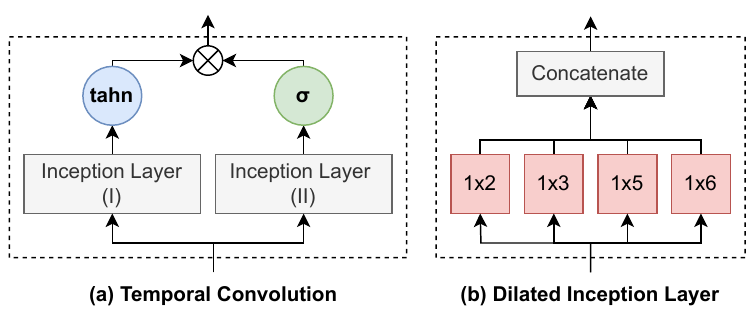}
  \end{center}
  \caption{Temporal convolution and dilated inception layer.}
  \label{img:tc}
\end{figure}

\paragraph{\textbf{Temporal Convolution Module}} For the weighted GNN forecasting component, the inputs are first transformed by a linear layer and then passed to the temporal convolution module. We employ the causal convolution~\cite{yu2015multi_causalconv} for our temporal convolution module. Causal convolution uses parallel computation to model temporal features with a non-recursive process. It applies 1D convolutional filters to extract high-level temporal context features. Setting different kernel sizes in causal convolution enables the mining of temporal patterns with various ranges in time series data. Unlike MTGNN~\cite{wu2020connecting_mtgnn}, we do not use dilated temporal convolution to expand the kernel size, as smaller window sizes are used in multivariate anomaly detection. Thus, CGAD uses a fixed kernel size. However, choosing the right kernel size is quite challenging. To balance the signal pattern effectively, the filter size must be carefully selected and not too large or too small.

To address that challenge, we keep the dilated inception layer~\cite{szegedy2015going_inception} as in the MTGNN model and use multiple filter sizes to maximize the performance of the causal convolution network. In forecast-based anomaly detection, it has been demonstrated that small input sequence lengths result in better performance for forecasting one value ahead~\cite{deng2021graph_GDN}. Since the granularity of measurement for real-time anomaly detection is commonly represented in units such as seconds, minutes, or hours, it makes 60 is a fundamental number. In contrast to the MTGNN model~\cite{wu2020connecting_mtgnn}, we selected filter sizes based on the smallest common divisors of 60, such as 2, 3, 5, and 6. This choice allows filter sizes of $1 \times 2$, $1 \times 3$, $1 \times 5$, and $1 \times 6$ to cover the common input sequences of time series in the anomaly detection problem.

The high-level architecture of the temporal convolution module is shown in Fig.~\ref{img:tc}. The time series input $\textbf{\textit{x}}\in \mathbb{R}^T$ is transformed using a linear layer before entering the temporal convolution module to produce a vector sequence $\textbf{\textit{z}}\in  \mathbb{R}^D$. The vector sequence $\textbf{\textit{z}}$ is then processed into two different inception layers. Given four filters, $\textbf{\textit{f}}_{1\times 2}\in  \mathbb{R}^2$, $\textbf{\textit{f}}_{1\times 3}\in  \mathbb{R}^3$, $\textbf{\textit{f}}_{1\times 5}\in  \mathbb{R}^5$, and $\textbf{\textit{f}}_{1\times 6}\in \mathbb{R}^6$ to be used as causal convolution process, the inception layer calculation will be:
\begin{equation}
  \label{eq:inception}
  \textbf{\textit{z}} = \textup{concat}(\textbf{\textit{z}}\star \textbf{\textit{f}}_{1\times 2},\textbf{\textit{z}}\star \textbf{\textit{f}}_{1\times 3},\textbf{\textit{z}}\star \textbf{\textit{f}}_{1\times 5},\textbf{\textit{z}}\star \textbf{\textit{f}}_{1\times 6}),
\end{equation}
where $\star$ represents the convolutional operator. Thus, the causal convolution represented by $\textbf{\textit{z}}\star f_{1 \times k}$ can be expressed by:
\begin{equation}
  \label{eq:cauconv}
  (\textbf{\textit{z}}\star \textbf{\textit{f}}_{1\times k}) (t) = \sum\limits_{s=0}^{k-1} f_{1\times k}(s) \cdot \textbf{\textit{z}}(t - s).
\end{equation}
In addition to the causal convolution process in inception layers, we need to consider the receptive field size of convolutional networks that grow in a linear progression based on the depth of the network and kernel size. By knowing the receptive field size, we are able to balance the input and output size of the convolutional operation by adding some control padding. The receptive field of the convolution network can be calculated as follows:
\begin{equation}
  \label{eq:receptive}
  r = m(k-1) + 1,
\end{equation}
where $r$ is the receptive field, $m$ is number of convolution layers, and $k$ is the kernel size. In the last step of temporal convolution, we use the gating mechanism~\cite{dauphin2017language_gatingmec}, which has proven to be powerful and aims to control information flow through layers in the temporal convolution process. There are two activation functions: the tangent hyperbolic as a filter and the sigmoid activation function to control the information that needs to be passed to the next module. Thus, the gating mechanism becomes:
\begin{equation}
  \label{eq:gatting}
  \textbf{\textit{h}} = \textup{tanh}(\mathbf{\theta}_1\star \textbf{\textit{z}} +b_1) \odot  \sigma(\mathbf{\theta}_2\star \textbf{\textit{z}} + b_2),
\end{equation}
where $z$ represent filtered sequence from inception layers, $\mathbf{\theta}_1$,$\mathbf{\theta}_2$, $b_1$ and $b_2$ are model parameters, $\odot$ is the element-wise product, $tanh$ denotes a tangent hyperbolic activation function, and $\sigma$ is the sigmoid activation function. Next, the result of gated TCN is passed to the GCN module and the skip connection layer.
\begin{figure}[h]
  \begin{center}
    \includegraphics[width=0.425\textwidth]{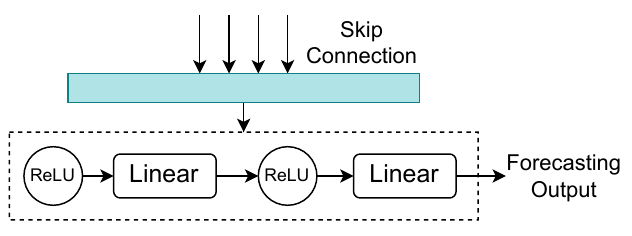}
  \end{center}
  \caption{Skip connection and output module.}
  \label{img:out}
  \vskip-5pt
\end{figure}

\paragraph{\textbf{Skip Connection and Output Module}} A skip connection module collects the results of the temporal convolution module in every layer to prevent the temporal information from succumbing to vanishing gradient problems. As shown in Fig.~\ref{img:out}, skip connection layers are essentially $1 \times L_i$ standard convolution, where $L_i$ is the sequence length of the inputs to the $i^{th}$ skip connection layers. Then, all results of the skip connection layers are summed together before passing to the output module. The output module consists of two linear layers to transform the channel dimension to the desired output dimension to forecast one-time step ahead for all nodes of the time series data.

We use a mean squared error (MSE) training objective between the predicted output and observed data. The MSE loss function can be calculated as follows:
\begin{equation}
  \label{eq:mse}
  \textup{loss}_{\mathrm{MSE}} = \frac{1}{T - w} \sum\limits_{t=w+1}^{T}\| \hat{\textbf{\textit{x}}}_{:,t} - \textbf{\textit{x}}_{:,t}\|_2^2,
\end{equation}
where $T$ is the maximum observation time in multivariate time series data, $w$ is the window size, $\hat{\textbf{\textit{x}}}_{:,t} \in \mathbb{R}^N$ is the result of single step forecasting for all nodes, and $\textbf{\textit{x}}_{:,t} \in \mathbb{R}^N$ is observed data for all nodes at a specific time step $t$. Furthermore, we summarize the weighted GNN forecasting training process in Algorithm~\ref{alg:2}.

\begin{algorithm}[tb]
\small
  \centering
  \caption{Training Process of CGAD Weighted GNN Forecasting Model}
  \label{alg:2}
  \begin{algorithmic}[1]
    \REQUIRE Multivariate time series data $X$, causal graph adjacency matrix $A$, batch size $\varDelta$, learning rate $\gamma$, window size $w$, initialized weighted GNN forecasting model $f(\cdot)$ with parameters $\Theta$
    \ENSURE Trained model parameters $\Theta$
    \WHILE {Not Converged}
      \STATE $\mathcal{X} \leftarrow \text{Sample a batch of input sequences from } X \in \mathbb{R}^{\varDelta \times N \times w}$
      \STATE $\mathcal{Y} \leftarrow \text{Sample corresponding target values from } X \in \mathbb{R}^{\varDelta \times N}$
      \STATE $\hat{\mathcal{Y}} \leftarrow f(\mathcal{X}, A, \Theta)$ \COMMENT{Predict target values using the model}
      \STATE $\textup{loss}_{\mathrm{MSE}} \leftarrow \textup{MSE}(\hat{\mathcal{Y}}, \mathcal{Y})$ \COMMENT{Compute MSE loss}
      \STATE $\nabla_{\Theta} \leftarrow \text{Compute gradients of } \Theta \text{ w.r.t. the loss}$
      \STATE $\Theta \leftarrow \Theta - \gamma \cdot \nabla_{\Theta}$ \COMMENT{Update model parameters using learning rate $\gamma$}
    \ENDWHILE 
  \end{algorithmic}
\end{algorithm}

\subsection{Median Deviation Scoring}
The developed forecasting model is implemented to test data containing anomalous samples. A threshold-based approach is used, whereby the observation at time $t$ is classed as an anomaly if its anomaly score exceeds the threshold. We first compare the forecast result with the actual observed value at time $t$ to calculate the anomaly score. The error at time $t$ for sensor $i$ can be computed as:
\begin{equation}
  \label{eq:err}
  \textup{error}_{i,t} = |\hat{\textbf{\textit{x}}}_{i,t} - \textbf{\textit{x}}_{i,t}|.
\end{equation}

After computing the error value, the anomaly score for sensor $i$ at time $t$ is standardized using the z-score. We chose the z-score over standard normalization techniques such as min-max because the z-score ensures that the scores are placed on a common scale and proves to be more robust when dealing with data anomalies. Specifically, we employ a modified z-score based on median absolute deviation (MAD), which is even more resistant to outliers than the traditional z-score~\cite{jamshidi2022detecting_mad}. In contrast to the traditional z-score, the modified z-score uses the median instead of the mean and median absolute deviation instead of the standard deviation. These median values, which replace the mean, ensure that anomalies do not influence the z-score calculation while assigning higher scores for anomalies far from 0. This robustness facilitates easier comparison and identification of outliers. The anomaly score for sensor $i$, $\textbf{\textit{a}}_{i,t}$, can be calculated in terms of the modified z-score:
\begin{equation}
  \label{eq:zscore}
  \textbf{\textit{a}}_{i,t} = \frac{\textup{error}_{i,t} - \textup{med}_i}{\textup{MAD}_i},
\end{equation}
where $\textup{med}_i$ represent the median of the $\textup{error}$ values of the specific time series $\textbf{\textit{x}}_{i,:}$. $\textup{MAD}_i = \textup{med}(|\textup{error}_{i,t} - \textup{med}_i|)$ which gives MAD values of the series $\textbf{\textit{x}}_{i,:}$. In this work, we consider an anomaly event as a global anomaly. Namely, if a single node raises an anomaly alarm at time step $t$, we consider it an anomaly event for the whole system. Thus, we need to aggregate all anomaly scores across different nodes. %
The collective anomaly score at a specific time $t$, denoted as $\textbf{\textit{s}}_{t}$, is calculated by aggregating the individual anomaly scores using the maximum function as follows:

\begin{equation}
  \label{eq:colscore}
  s_{t} = \underset{i \in N}{\textup{MAX}}(\textbf{\textit{a}}_{i,t}),
\end{equation}
where $\textup{MAX}$ is an aggregation function to get a collective anomaly score by selecting the highest value among all node anomaly scores at time $t$. Once we get that score, determining whether an event is an anomaly or a normal event can be described with this scenario:
\begin{equation}
  \label{eq:anom}
  y_{t} =
  \begin{cases}
    1 & \text{if $s_{t} > \tau ,$} \\
    0 & \text{otherwise},
  \end{cases}
\end{equation}
where $\tau$ denotes the anomaly threshold; if the collective anomaly score exceeds the threshold, the event at $t$ is classified as an anomaly. In this work, we use the peak over the threshold (POT)~\cite{siffer2017anomaly_pot} strategy to automate the threshold selection process, which utilizes extreme value theory based on the generalized Pareto distribution. Using that strategy enables the CGAD framework to identify suitable threshold values dynamically at a specific risk parameter. Ultimately, we summarize the CGAD anomaly detection process in Algorithm~\ref{alg:3}.

\begin{algorithm}[tb]
\small
  \centering
  \caption{Anomaly Detection Process}
  \label{alg:3}
  \begin{algorithmic}[1]
    \REQUIRE Multivariate time series data $X$, causal graph adjacency matrix $A$, and a trained weighted GNN forecasting model $f(\cdot)$ with parameters $\Theta$
    \ENSURE Anomaly prediction results
    \FOR{$t = 1$ to $T - 1$}
      \STATE $\textbf{\textit{x}}_{i,t-w} \leftarrow \text{Generate input sequence from } X$
      \STATE $\textbf{\textit{x}}_{:,t} \leftarrow \text{Target values from } X$
      \STATE $\hat{\textbf{\textit{x}}}_{:,t} \leftarrow f(\hat{\textbf{\textit{x}}}_{i,t-w : t-1}, A, \Theta)$ \COMMENT{Forecast using the trained model}
      \STATE $\textup{error}_{i,t} \leftarrow |\hat{\textbf{\textit{x}}}_{i,t} - \textbf{\textit{x}}_{i,t}|$ \COMMENT{Compute prediction error}
      \STATE $\textbf{\textit{a}}_{i,t} \leftarrow$ Calculate node-level anomaly scores \COMMENT{Refer to Equation~\ref{eq:zscore}}
      \STATE $\textbf{\textit{a}}_{i,t} \leftarrow$ Aggregate node-level scores $\textbf{\textit{s}}_{t}$ into collective score \COMMENT{Refer to Equation~\ref{eq:colscore}}
      \STATE $\textbf{\textit{y}}_{t} \leftarrow$ Detect anomalies using the collective score \COMMENT{Refer to Equation~\ref{eq:anom}}
    \ENDFOR
  \end{algorithmic}
\end{algorithm}

\subsection{Complexity Analysis}
CGAD starts with the causal graph generation, where calculating TE between pairs of time series has a complexity of $O(w)$ per calculation, where $w$ is the window size. For $N$ nodes, there are $N(N-1)$ pairwise comparisons, excluding self-loops, resulting in a complexity of $O(N^2 \times w)$. With $G$ samples, the overall complexity of graph generation becomes $O(G \times N^2 \times w)$. The weighted GNN forecasting module has a graph convolution complexity of $O(N^2 \times F)$, where $F$ is the hidden feature dimension, and a temporal convolution complexity of $O(T \times m)$, where $T$ is the sequence length and $m$ is the number of layers, resulting in a total complexity of $O(N^2 \times F) + O(T \times m)$. Lastly, the median deviation scoring involves sorting for median calculation, which, for each node $i$, takes $O(T \log T)$. Since there are 
$N$ nodes, the overall complexity of the median deviation scoring is $O(N \times T \log T)$.

\section{Experiments}
We conducted extensive experiments on multivariate time series anomaly detection, aiming to answer the following four research questions:

\begin{itemize}
    \item \textbf{RQ1}: Does CGAD outperform baseline methods for anomaly detection on multivariate time series data?
    \item \textbf{RQ2}: Do various components in CGAD contribute to the overall model performance?
    \item \textbf{RQ3}: How does the causal graph generation technique in CGAD improve the flexibility of the graph structures compared to previous techniques?
    \item \textbf{RQ4}: How can the anomaly detection process in CGAD be diagnosed across nodes in multivariate time series data?

\end{itemize}

\subsection{Datasets}
We conduct extensive experiments on five real-world datasets from various domains in cyber-physical systems such as The Secure Water Treatment (SWAT)~\cite{mathur2016swat}, Water Distribution (WADI)~\cite{ahmed2017wadi}, Soil Moisture Active Passive (SMAP)~\cite{hundman2018detecting_lstmndt}, Mars Science Laboratory rover (MSL)~\cite{hundman2018detecting_lstmndt}, and Server Machine Dataset (SMD)~\cite{su2019robust_omnianomaly}. The datasets are selected to represent multivariate time series anomaly detection scenarios constructed from multiple sensors capturing time series data. These datasets are chosen from various fields, including industrial control systems, telemetry spacecraft, and computer internet networks, to assess the model's ability to perform detection across different domains. The following are the descriptions of the datasets:

\begin{itemize}
    \item \textbf{The Secure Water Treatment (SWAT)}~\cite{mathur2016swat}. SWAT is collected from a water treatment test bed, integrating the digital and physical elements to control and monitor system behavior. It has 51 operated sensors for 11 days, with 7 days of normal and 4 days of attack scenarios.  
    \item \textbf{Water Distribution (WADI)}~\cite{ahmed2017wadi}. WADI is an extension experiment of SWAT, consisting of 127 sensors and actuators simulating physical attacks with 14 days of normal and 2 days of attack scenarios.
    \item \textbf{Soil Moisture Active Passive (SMAP)}~\cite{hundman2018detecting_lstmndt}.The SMAP dataset is derived from spacecraft telemetry signals provided by NASA. 
    \item \textbf{Mars Science Laboratory rover (MSL)}~\cite{hundman2018detecting_lstmndt}. Similar to the SMAP dataset, MSL is a spacecraft dataset collected by NASA. 
    \item \textbf{Server Machine Dataset (SMD)}~\cite{su2019robust_omnianomaly}. SMD is collected in the OmniAnomaly work from a large Internet company, which includes 5 weeks of stack trace data on resource utilization with 38 sensors for each entity.
    \item \textbf{Pooled Server Metrics (PSM)}~\cite{abdulaal2021practical_PSM}. PSM consists of diverse application server node data from eBay, providing 25 dimensions of multivariate time series. The dataset is collected with 13 weeks of training and 8 weeks of testing data.
    
\end{itemize}

The statistics of each dataset are in Table~\ref{tb:datasets}, including the size of the train and test datasets, the number of nodes (i.e., number of sensors), the number of subsets used from each dataset, and the percentage of observations that are anomalies. Moreover, for building the forecasting model, we split each training dataset again, which contains normal sequences without anomaly, into two: actual training sets (80\%) to train the forecasting model and validation sets (20\%) to help the model selection.

\begin{table}[!hb]
  \centering
  \small
\caption{Basic statistics of datasets.}
  \label{tb:datasets}
    \resizebox{0.5\textwidth}{!}{\begin{tabular}{ccccc}
      \hline
      \multicolumn{1}{c}{Datasets} & \multicolumn{1}{c}{Train} & \multicolumn{1}{c}{Test} & \multicolumn{1}{c}{Nodes (Subsets)} & \multicolumn{1}{c}{Anomalies (\%)} \\ \hline
      SWAT & 496800 & 449919 & 55 (1) & 11.98 \\
      WADI & 1048571 & 172801 & 123 (1) & 5.71 \\
      SMAP & 135183 & 427617 & 25 (55) & 13.13 \\
      MSL & 58317 & 73729 & 55 (27) & 10.72 \\
      SMD & 708405 & 708420 & 38 (28) & 5.37 \\ 
      PSM & 132481 & 87841 & 25 (1) & 27.76 \\\hline
      \end{tabular}}
\end{table}

\subsection{Baseline Methods}
We compare our proposed model's performance with various state-of-the-art multivariate time series anomaly detection frameworks. These models include both non-graph-based and graph-based methods. The following are the details on each baseline method:

\begin{itemize}
    \item \textbf{LSTM-NDT}~\cite{hundman2018detecting_lstmndt}. The work combines LSTMs with VAE to build a reconstruction-based model and uses nonparametric dynamic thresholding (NDT) to detect anomalies.
    \item \textbf{DAGMM}~\cite {zong2018deep_DAGMM}. DAGMM uses a deep autoencoder and reconstruction error for each input data point, which is then fed into a Gaussian Mixture Model (GMM) for processing.
    \item \textbf{OmniAnomaly}~\cite{su2019robust_omnianomaly}. OmniAnomaly learns a representation of the normal patterns through stochastic variable connection and planar normalizing flow, reconstructs the input data by the representation, and uses reconstruction probabilities to determine anomalies.
    \item \textbf{USAD}~\cite{audibert2020usad}. USAD utilizes auto-encoders with two decoders in an adversarial training framework to build a reconstruction model, with hyperparameters introduced to balance false positives and true positives.
    \item \textbf{MAD-GAN}~\cite{li2019mad_madgan}. MAD-GAN adopts an LSTM-based GAN model to establish the multivariate correlations among time series data and utilizes discrimination and reconstruction losses to detect anomalies.
    \item \textbf{TranAD}~\cite{tuli2022tranad}. TranAD uses transformer-based encoder-decoder structures and self-conditioning with adversarial training to amplify reconstruction errors for anomaly detection.
    \item \textbf{DAEMON}~\cite{chen2023adversarial_DAEMON}. DAEMON uses two discriminators to adversarially train an autoencoder to learn normal patterns in multivariate time series data during training. It uses the reconstruction error to detect anomalies.
    \item \textbf{MEMTO}~\cite{song2024memto}. The work addresses over-generalization issues in reconstruction-based models using a gated memory module and a bi-dimensional deviation detection criterion.
    \item \textbf{MTAD-GAT}~\cite{zhao2020multivariate_mtadgat}. MTAD-GAT is a graph-based method that utilizes a feature-oriented GAT layer to learn correlations between time series and a GRU network to capture sequential patterns. An anomaly score is derived from the prediction and reconstruction loss.
    \item \textbf{GDN}~\cite{deng2021graph_GDN}. GDN is a graph-based model that builds a graph structure by computing cosine similarity between variables and employs an attention-based technique to predict future values of multivariate time series data with graph deviation scores to detect anomalies.
    \item \textbf{GTA}~\cite{chen2021learning_gta}. GTA is a graph-based technique that utilizes the Gumbel-softmax to learn the connections between variables and employs a forecasting-based strategy with temporal convolution and attention network to detect anomalies.
    \item \textbf{DVGCRN}~\cite{chen2022deep_DVGCRN}.  DVGCRN is a graph-based technique that uses graph convolutional and recurrent structures to capture spatial-temporal correlations and hierarchical relationships.

\end{itemize}

We use publicly available code for all baseline methods and adopt their reported parameter settings.

\subsection{Experimental Setup}
\label{ap:exp}
We implemented our method using Python version 3.9.12, Pytorch library version 1.13.1 with CUDA 11.6, and Pytorch Geometric library version 2.2.0. The model was trained by Adam optimizer with a learning rate $1 \times 10^{-3}$. The hardware used to run the experiment was AMD Ryzen 7 5800H @ 3.20 GHz with NVIDIA GeForce RTX 3050 Ti Laptop GPU.

For single-step forecasting, 3 graph convolution modules and 3 temporal convolution modules were used. Similar to the MTGNN model, the initial $1 \times 1$ convolution has 1 input and 16 output channels. The skip connection layers were all set to 32 output channels. The input and output of the gating mechanism ($\textup{tanh}$ and $\sigma$) in the temporal convolution were set with the same size, 16, and a dilation factor of 1. The two-layer GCN model has 16 input and output channels and 32 hidden layer channels. Additionally, the first layer and the second layer of output modules were set to 64 and 1, respectively. The common first data pre-processing stage is implemented, where the data is normalized and split into time windows with 15 window sizes and 1 single-step output. We trained the model with 10 epochs, and the batch size was set to 32 for all datasets. We repeated our model with five runs to show consistency in performance.

\begin{table}[!h]
\tiny
\centering
\caption{Experiment results. \textbf{Bold} indicates the best performance; \underline{underline} denotes the second-best.}
\label{tb:results}
\resizebox{0.8\textwidth}{!}{
\begin{tabular}{llllllllll}
\hline
\multicolumn{1}{c|}{\multirow{2}{*}{Methods}} & \multicolumn{3}{c|}{SWAT} & \multicolumn{3}{c|}{WADI} & \multicolumn{3}{c}{SMAP} \\ \cline{2-10} 
\multicolumn{1}{c|}{} & \multicolumn{1}{c}{F1} & \multicolumn{1}{c}{$\textup{F1}_c$} & \multicolumn{1}{c|}{$\textup{F1}_{PA}$} & \multicolumn{1}{c}{F1} & \multicolumn{1}{c}{$\textup{F1}_c$} & \multicolumn{1}{c|}{$\textup{F1}_{PA}$} & \multicolumn{1}{c}{F1} & \multicolumn{1}{c}{$\textup{F1}_c$} & \multicolumn{1}{c}{$\textup{F1}_{PA}$} \\ \hline
\multicolumn{10}{c}{\cellcolor[HTML]{E8E8E8}Non Graph-based} \\ \hline
\multicolumn{1}{l|}{LSTM-NDT} & 0.7237 & 0.2400 & \multicolumn{1}{l|}{0.7116} & 0.2501
 & 0.4615 & \multicolumn{1}{l|}{0.5662} & \underline{0.4834} & \underline{0.5509} & 0.8767 \\
\multicolumn{1}{l|}{DAGMM} & 0.5508 & 0.1523 & \multicolumn{1}{l|}{0.8133} & 0.2948 & 0.2663 & \multicolumn{1}{l|}{0.1411} & 0.3333 & 0.2228 & 0.7752 \\
\multicolumn{1}{l|}{OmniAnomaly} & 0.7827 & 0.2558 & \multicolumn{1}{l|}{0.8391} & 0.2231 & 0.2665 & \multicolumn{1}{l|}{0.4577} & 0.2276 & 0.3750 & 0.8389 \\
\multicolumn{1}{l|}{USAD} & \underline{0.7922} & 0.5509 & \multicolumn{1}{l|}{0.8366} & 0.2322 & 0.1428 & \multicolumn{1}{l|}{0.4298} & 0.2280 & 0.3136 & 0.8637 \\
\multicolumn{1}{l|}{MAD-GAN} & 0.7477 & 0.5516 & \multicolumn{1}{l|}{0.8070} & 0.3701 & 0.5185 & \multicolumn{1}{l|}{0.3586} & 0.4194 & 0.4506 & 0.8134 \\
\multicolumn{1}{l|}{TranAD} & 0.6754 & 0.2012 & \multicolumn{1}{l|}{0.8155} & 0.2526 & 0.4672 & \multicolumn{1}{l|}{0.4951} & 0.4435 & 0.3285 & 0.8914 \\
\multicolumn{1}{l|}{DAEMON} & 0.2140 & 0.2411 & \multicolumn{1}{l|}{0.9476} & 0.1285 & 0.2114 & \multicolumn{1}{l|}{0.8906} & 0.3233 & 0.3657 & 0.9106 \\
\multicolumn{1}{l|}{MEMTO} & 0.6667 & \underline{0.8702} & \multicolumn{1}{l|}{\underline{0.9583}} & 0.1910 & 0.2503 & \multicolumn{1}{l|}{0.9045} & 0.4324 & 0.4405 & \textbf{0.9661} \\ \hline
\multicolumn{10}{c}{\cellcolor[HTML]{E8E8E8}Graph-based} \\ \hline
\multicolumn{1}{l|}{MTAD-GAT} & 0.7365 & 0.4687 & \multicolumn{1}{l|}{0.8114} & \underline{0.4371} & 0.5574 & \multicolumn{1}{l|}{0.4168} & 0.4471 & 0.4380 & 0.9017 \\
\multicolumn{1}{l|}{GDN} & \textbf{0.8087} & 0.4539 & \multicolumn{1}{l|}{0.9449} & \textbf{0.5701} & \underline{0.7640} & \multicolumn{1}{l|}{0.5695} & 0.4305 & 0.4188 & 0.8116 \\
\multicolumn{1}{l|}{GTA} & 0.7140 & 0.4091 & \multicolumn{1}{l|}{0.9099} & 0.2533 & 0.6499 & \multicolumn{1}{l|}{0.8781} & 0.2309 & 0.2698 & 0.9379 \\
\multicolumn{1}{l|}{DVGCRN} & 0.7633 & 0.6602 & \multicolumn{1}{l|}{0.9270} & 0.3297 & 0.6009 & \multicolumn{1}{l|}{\underline{0.9260}} & 0.4537 & 0.4642 & 0.9433 \\
\multicolumn{1}{l|}{CGAD} & 0.7518 & \textbf{0.8968} & \multicolumn{1}{l|}{\textbf{0.9611}} & 0.3727 & \textbf{0.7897} & \multicolumn{1}{l|}{\textbf{0.9488}} & \textbf{0.4994} & \textbf{0.5669} & \underline{0.9468} \\
 &  &  &  &  &  &  &  &  &  \\ \hline
\multicolumn{1}{c|}{\multirow{2}{*}{Methods}} & \multicolumn{3}{c|}{MSL} & \multicolumn{3}{c|}{SMD} & \multicolumn{3}{c}{PSM} \\ \cline{2-10} 
\multicolumn{1}{c|}{} & \multicolumn{1}{c}{F1} & \multicolumn{1}{c}{$\textup{F1}_c$} & \multicolumn{1}{c|}{$\textup{F1}_{PA}$} & \multicolumn{1}{c}{F1} & \multicolumn{1}{c}{$\textup{F1}_c$} & \multicolumn{1}{c|}{$\textup{F1}_{PA}$} & \multicolumn{1}{c}{F1} & \multicolumn{1}{c}{$\textup{F1}_c$} & \multicolumn{1}{c}{$\textup{F1}_{PA}$} \\ \hline
\multicolumn{10}{c}{\cellcolor[HTML]{E8E8E8}Non Graph-based} \\ \hline
\multicolumn{1}{l|}{LSTM-NDT} & \underline{0.4136} & 0.3234 & \multicolumn{1}{l|}{0.8087} & 0.2829 & 0.2625 & \multicolumn{1}{l|}{0.9047} & 0.4834 & 0.4956 & 0.9171 \\
\multicolumn{1}{l|}{DAGMM} & 0.1990 & 0.1947 & \multicolumn{1}{l|}{0.7007} & 0.2381 & 0.2588 & \multicolumn{1}{l|}{0.7096} & 0.4794 & 0.5042 & 0.9364 \\
\multicolumn{1}{l|}{OmniAnomaly} & 0.1909 & 0.3871 & \multicolumn{1}{l|}{0.8873} & \underline{0.4740} & 0.6632 & \multicolumn{1}{l|}{0.8855} & 0.4556 & 0.5010 & 0.8794 \\
\multicolumn{1}{l|}{USAD} & 0.2110 & 0.4385 & \multicolumn{1}{l|}{0.9108} & 0.4261 & 0.3138 & \multicolumn{1}{l|}{0.8616} & 0.4750 & 0.5238 & 0.9463 \\
\multicolumn{1}{l|}{MAD-GAN} & 0.3893 & 0.4053 & \multicolumn{1}{l|}{0.8751} & 0.1984 & 0.2003 & \multicolumn{1}{l|}{0.9155} & 0.4623 & 0.5415 & 0.9447 \\
\multicolumn{1}{l|}{TranAD} & 0.4007 & 0.4178 & \multicolumn{1}{l|}{0.9493} & 0.3145 & 0.3752 & \multicolumn{1}{l|}{0.9609} & \textbf{0.4940} & 0.4907 & 0.9112 \\
\multicolumn{1}{l|}{DAEMON} & 0.2835 & 0.3003 & \multicolumn{1}{l|}{\underline{0.9527}} & 0.2145 & \underline{0.6909} & \multicolumn{1}{l|}{\underline{0.9630}} & 0.4350 & 0.5817 & 0.9823 \\
\multicolumn{1}{l|}{MEMTO} & 0.4005 & \underline{0.5512} & \multicolumn{1}{l|}{0.9436} & 0.1697 & 0.1679 & \multicolumn{1}{l|}{0.9354} & 0.4494 & \underline{0.7894} & \underline{0.9838} \\ \hline
\multicolumn{10}{c}{\cellcolor[HTML]{E8E8E8}Graph-based} \\ \hline
\multicolumn{1}{l|}{MTAD-GAT} & 0.2749 & 0.3193 & \multicolumn{1}{l|}{0.9087} & 0.4000 & 0.3997 & \multicolumn{1}{l|}{0.8687} & 0.4605 & 0.4591 & 0.8563 \\
\multicolumn{1}{l|}{GDN} & 0.3502 & 0.4140 & \multicolumn{1}{l|}{0.8526} & 0.1884 & 0.1885 & \multicolumn{1}{l|}{0.6532} & 0.4591 & 0.4552 & 0.9157 \\
\multicolumn{1}{l|}{GTA} & 0.2179 & 0.2198 & \multicolumn{1}{l|}{0.9037} & 0.3510 & 0.5689 & \multicolumn{1}{l|}{0.9380} & 0.4333 & 0.4334 & 0.9810 \\
\multicolumn{1}{l|}{DVGCRN} & 0.3804 & 0.3878 & \multicolumn{1}{l|}{0.9141} &  0.3007 & 0.3207 & \multicolumn{1}{l|}{0.9157} & \underline{0.4934} & 0.5665 & 0.9474 \\
\multicolumn{1}{l|}{CGAD} & \textbf{0.4197} & \textbf{0.5839} & \multicolumn{1}{l|}{\textbf{0.9618}} & \textbf{0.4867} & \textbf{0.8177} & \multicolumn{1}{l|}{\textbf{0.9724}} & 0.4387 & \textbf{0.7935} & \textbf{0.9898} \\
 &  &  &  &  &  &  &  &  &  \\ \cline{1-4}
\multicolumn{1}{c|}{\multirow{2}{*}{Methods}} & \multicolumn{3}{c}{\textbf{Average Performance}} &  &  &  &  &  &  \\ \cline{2-4}
\multicolumn{1}{c|}{} & \multicolumn{1}{c}{F1} & \multicolumn{1}{c}{$\textup{F1}_c$} & \multicolumn{1}{c}{$\textup{F1}_{PA}$} &  &  &  &  &  &  \\ \cline{1-4}
\multicolumn{4}{c}{\cellcolor[HTML]{E8E8E8}Non Graph-based} &  &  &  &  &  &  \\ \cline{1-4}
\multicolumn{1}{l|}{LSTM-NDT} & 0.4395 & 0.3890 & 0.7975 &  &  &  &  &  &  \\
\multicolumn{1}{l|}{DAGMM} & 0.3492 & 0.2665 & 0.6794 &  &  &  &  &  &  \\
\multicolumn{1}{l|}{OmniAnomaly} & 0.3923 &  0.4081 &  0.7980 &  &  &  &  &  &  \\
\multicolumn{1}{l|}{USAD} & 0.3941 & 0.3806 & 0.8081 &  &  &  &  &  &  \\
\multicolumn{1}{l|}{MAD-GAN} & 0.4312 & 0.4446 & 0.7857 &  &  &  &  &  &  \\
\multicolumn{1}{l|}{TranAD} & 0.4301 & 0.3801 & 0.8372 &  &  &  &  &  &  \\
\multicolumn{1}{l|}{DAEMON} & 0.2665 & 0.3985 & 0.9411 &  &  &  &  &  &  \\
\multicolumn{1}{l|}{MEMTO} & 0.3850 & \underline{0.5116} & \underline{0.9486} &  &  &  &  &  &  \\ \cline{1-4}
\multicolumn{4}{c}{\cellcolor[HTML]{E8E8E8}Graph-based} & \multicolumn{1}{c}{} & \multicolumn{1}{c}{} & \multicolumn{1}{c}{} & \multicolumn{1}{c}{} & \multicolumn{1}{c}{} & \multicolumn{1}{c}{} \\ \cline{1-4}
\multicolumn{1}{l|}{MTAD-GAT} & 0.4594 & 0.4404 & 0.7939 &  &  &  &  &  &  \\
\multicolumn{1}{l|}{GDN} & \underline{0.4678} & 0.4491 & 0.7913 &  &  &  &  &  &  \\
\multicolumn{1}{l|}{GTA} & 0.3667 & 0.4252 & 0.9248 &  &  &  &  &  &  \\
\multicolumn{1}{l|}{DVGCRN} & 0.4535 & 0.5001 & 0.9289 &  &  &  &  &  &  \\
\multicolumn{1}{l|}{CGAD} & \textbf{0.4948} & \textbf{0.7414} & \textbf{0.9635} &  &  &  &  &  &
\end{tabular}}
\end{table}

\subsection{Evaluation Metrics}
With an increasing number of works on multivariate time series anomaly detection over the years, many evaluation protocols and metrics have been proposed~\cite{audibert2020usad,garg2021evaluation_composite,kim2022towards_rigorous}. The most straightforward protocol is to calculate point-wise F1 scores, which assume that each observation at each timestamp is independent. However, most anomalies in multivariate time series appear consecutively, forming anomaly segments. Thus, classical point-wise evaluation metrics may not adequately reflect the performance of continuous segments of anomaly events~\cite{garg2021evaluation_composite}.

Motivated to detect anomalies in continuous events or episodes, most works currently apply an evaluation protocol called the point-adjustment strategy ($\textup{F1}_{PA}$)~\cite{audibert2020usad, su2019robust_omnianomaly, tuli2022tranad, chen2021learning_gta}. For a better view, $\textup{F1}_{PA}$ works as follows: if at least one event in the contiguous anomaly segment that the model detects as an anomaly, the entire segment is also considered to be correctly predicted as an anomaly. However, some works suggest that PA can overestimate the detection performance~\cite{kim2022towards_rigorous}, indicating that presenting $\textup{F1}_{PA}$ alone is insufficient as an evaluation metric. Another work introduces an evaluation protocol called composite score ($\textup{F1}_{c}$)~\cite{garg2021evaluation_composite} that addresses the problem of overestimation in the PA strategy and offers an ideal detector by calculating the F1 score using event-wise recall and point-wise precision.

To intuitively and comprehensively understand the real effect of multivariate time series anomaly detection performance, we evaluated our model's and baseline models' anomaly detection capability using three metrics. The first metric is the typical point-wise score (F1). The second is the composite score ($\textup{F1}_c$), calculated as $\textup{F1}_c=\frac{2 \times P \times R_e}{P+R_e}$, where $P$ is point-wise precision, defined as $P = \frac{TP}{TP+FP}$, with TP and FP being the numbers of true positives and false positives based on point-to-point observations, and $\textup{R}_e$ is event-wise recall ($\textup{R}_e$), defined as $\textup{R}e = \frac{TP_e}{\textup{\#GT events}}$, where $TP_e$ is the number of true positives based on anomaly events, and \#GT events represents the total ground truth of anomaly events. The third metric is the point-adjustment score ($\textup{F1}_{PA}$). Consistent with previous works~\cite{tuli2022tranad, deng2021graph_GDN, su2019robust_omnianomaly}, we use the peak over threshold (POT) technique to automate threshold selection. A similar setting to obtain hyperparameters for POT based on a grid search to find the best possible threshold ensures a fair comparison.

\subsection{RQ1. Experimental Results}  
The anomaly detection results on six datasets are presented in Table~\ref{tb:results}. The average performance of all datasets shows that CGAD significantly improved the detection scores over state-of-the-art methods with a 9\% average improvement over all three different detection metrics. Specifically, CGAD achieves the best average performance scores of 0.4948, 0.7445, and 0.9635 for $\textup{F1}$, $\textup{F1}_c$, and $\textup{F1}_{PA}$, respectively.

Based on the typical point-wise score (F1), CGAD outperforms the baseline models on SMAP, MSL, and SMD datasets. GDN scores best in the SWAT and WADI datasets with F1 scores of 0.8087 and 0.5701, respectively. The higher point-wise score (F1) indicates better models for detecting anomalies in individual anomalous time points. However, point-wise evaluation is inadequate to guarantee better performance in detecting anomalies in continuous segments or events, measured with the second metric of the composite score ($\textup{F1}_c$). The composite score ($\textup{F1}_c$) shows that CGAD outperforms other models with the best scores over all datasets. It also shows that in the SWAT dataset, CGAD can double the performance compared to the second best $\textup{F1}_c$, which is MAD-GAN with $\textup{F1}_c$ of 0.5516. This indicates the strength of CGAD in detecting anomalies at the event level, which is more important than individual anomalous time points. Additionally, using the popular metric point-adjust trick ($\textup{F1}_{PA}$), CGAD is also significantly better than other algorithms.

The overall trend of CGAD and across baseline methods shows lower performance in the point-wise score ($\textup{F1}$) compared to the composite score ($\textup{F1}_{c}$) and point-adjusted F1 score ($\textup{F1}_{PA}$). These findings suggest that point-wise anomaly detection, which focuses on classifying individual data points, is more challenging because it can be easily confused with noise, making it difficult to distinguish true anomalies. In contrast, event-wise anomalies, which focus on continuous detections over a certain period, provide more robust and clearer patterns, making them easier for models to detect. Detecting anomaly events is more reliable as real-world anomalies are not just one-off points but rather behavior patterns over an extended period that indicate underlying problems. Detecting longer events like event-wise anomalies provides more actionable insights that are more relevant for maintenance, troubleshooting, or decision-making.

CGAD consistently outperforms existing algorithms due to its superior ability to model meaningful relationships in multivariate time series data, enhancing anomaly detection performance. Unlike non-graph methods, which typically model the temporal patterns of each variable independently, CGAD uses a graph structure based on transfer entropy to capture causal relationships between variables. This approach overcomes limitations of other graph-based methods: (i) MTAD-GAN assumes an unrealistic, fully connected graph, with edges connecting all nodes/variables, (ii) GDN's top-$k$ strategy results in an inflexible uniform degree distribution for each node, (iii) GTA's end-to-end learning, leveraging the Gumbel-softmax, disregards domain knowledge, making it more susceptible to overfitting, (iv) DVGCRN is designed to handle noisy data, but it doesn't accurately capture the relationships between sensors. 
CGAD's causal graph integrates domain knowledge, providing a more meaningful and flexible structure, without enforcing uniformity. By combining this improved spatial analysis with temporal analysis via a GNN and TCN, CGAD excels in predicting future values in multivariate time series datasets and detecting anomalies in multivariate time series data. Furthermore, CGAD's median deviation scoring method offers more robust anomaly scoring, providing clearer distinctions between normal data points and anomalies, even in noisy datasets. 

There are potential challenges that could arise in multivariate time series anomaly detection tasks. One such challenge is the presence of distribution shifts in normal patterns or the emergence of unseen patterns in the training data, which could affect the generalization capabilities of the framework~\cite{han2023anomaly_normal}. More studies to address this aspect within the context of CGAD are left as future work. In addition, we can incorporate continuous or lifelong learning mechanisms~\cite{febrinanto2023graph}. This would enable CGAD to dynamically adapt its anomaly scoring to new information and evolving anomaly scenarios, eliminating the need for retraining processes.

\begin{table}[!ht]
\centering
\small
\caption{Ablation study results. \textbf{Bold} indicates the best performance.}
\label{tb:ablation}
\resizebox{0.7\textwidth}{!}{
\begin{tabular}{llllllllll}
\hline
\multicolumn{1}{l|}{\multirow{2}{*}{Methods}} & \multicolumn{3}{c|}{SWAT} & \multicolumn{3}{c|}{WADI} & \multicolumn{3}{c}{SMAP} \\ \cline{2-10} 
\multicolumn{1}{l|}{} & \multicolumn{1}{c}{F1} & \multicolumn{1}{c}{$\textup{F1}_c$} & \multicolumn{1}{c|}{$\textup{F1}_{PA}$} & \multicolumn{1}{c}{F1} & \multicolumn{1}{c}{$\textup{F1}_c$} & \multicolumn{1}{c|}{$\textup{F1}_{PA}$} & \multicolumn{1}{c}{F1} & \multicolumn{1}{c}{$\textup{F1}_c$} & \multicolumn{1}{c}{$\textup{F1}_{PA}$} \\ \hline
\multicolumn{1}{l|}{CGAD} & \textbf{0.7518} & \textbf{0.8968} & \multicolumn{1}{l|}{\textbf{0.9611}} & \textbf{0.3727} & \textbf{0.7897} & \multicolumn{1}{l|}{\textbf{0.9488}} & \textbf{0.4994} & \textbf{0.5669} & \textbf{0.9468} \\ \hline
\multicolumn{1}{l|}{\textit{- Caugraph}} & 0.6994 & 0.8494 & \multicolumn{1}{l|}{0.9354} & 0.2904 & 0.7582 & \multicolumn{1}{l|}{0.9438} & 0.4087 & 0.4790 & 0.9128 \\
\multicolumn{1}{l|}{\textit{- GConv}} & 0.6753 & 0.7945 & \multicolumn{1}{l|}{0.9392} & 0.3571 & 0.7597 & \multicolumn{1}{l|}{0.9430} & 0.4309 & 0.5251 & 0.9244 \\
\multicolumn{1}{l|}{- Zscore} & 0.2254 & 0.4164 & \multicolumn{1}{l|}{0.8210} & 0.2565 & 0.6141 & \multicolumn{1}{l|}{0.8842} & 0.4287 & 0.4449 & 0.9164 \\
 &  &  &  &  &  &  &  &  &  \\ \hline
\multicolumn{1}{l|}{\multirow{2}{*}{Methods}} & \multicolumn{3}{c|}{MSL} & \multicolumn{3}{c|}{SMD} & \multicolumn{3}{c}{PSM} \\ \cline{2-10} 
\multicolumn{1}{l|}{} & \multicolumn{1}{c}{F1} & \multicolumn{1}{c}{$\textup{F1}_c$} & \multicolumn{1}{c|}{$\textup{F1}_{PA}$} & \multicolumn{1}{c}{F1} & \multicolumn{1}{c}{$\textup{F1}_c$} & \multicolumn{1}{c|}{$\textup{F1}_{PA}$} & \multicolumn{1}{c}{F1} & \multicolumn{1}{c}{$\textup{F1}_c$} & \multicolumn{1}{c}{$\textup{F1}_{PA}$} \\ \hline
\multicolumn{1}{l|}{CGAD} & \textbf{0.4197} & \textbf{0.5839} & \multicolumn{1}{l|}{\textbf{0.9618}} & \textbf{0.4867} & \textbf{0.8177} & \multicolumn{1}{l|}{\textbf{0.9724}} & \textbf{0.4383} & \textbf{0.7935} & \textbf{0.9898} \\ \hline
\multicolumn{1}{l|}{\textit{- Caugraph}} & 0.3550 & 0.2434 & \multicolumn{1}{l|}{0.8644} & 0.2670 & 0.3975 & \multicolumn{1}{l|}{0.9656} & 0.4379 & 0.7811 & 0.9843 \\
\multicolumn{1}{l|}{\textit{- GConv}} & 0.3407 & 0.2550 & \multicolumn{1}{l|}{0.8227} & 0.3072 & 0.3619 & \multicolumn{1}{l|}{0.9360} & 0.4346 & 0.6793 & 0.9365 \\
\multicolumn{1}{l|}{\textit{- Zscore}} & 0.3222 & 0.3286 & \multicolumn{1}{l|}{0.8195} & 0.1566 & 0.2700 & \multicolumn{1}{l|}{0.9581} & 0.4349 & 0.6722 & 0.9680 \\
 &  &  &  &  &  &  &  &  &  \\ \cline{1-4}
\multicolumn{1}{l|}{\multirow{2}{*}{Methods}} & \multicolumn{3}{c}{\textbf{Average Performance}} &  &  &  &  &  &  \\ \cline{2-4}
\multicolumn{1}{l|}{} & \multicolumn{1}{c}{F1} & \multicolumn{1}{c}{$\textup{F1}_c$} & \multicolumn{1}{c}{$\textup{F1}_{PA}$} &  &  &  &  &  &  \\ \cline{1-4}
\multicolumn{1}{l|}{CGAD} & \textbf{0.4948} & \textbf{0.7414} & \textbf{0.9635} &  &  &  &  &  &  \\ \cline{1-4}
\multicolumn{1}{l|}{\textit{- Caugraph}} & 0.4097 & 0.5848 & 0.9344 &  &  &  &  &  &  \\
\multicolumn{1}{l|}{\textit{- GConv}} & 0.4243 & 0.5626 & 0.9170 &  &  &  &  &  &  \\
\multicolumn{1}{l|}{\textit{- Zscore}} & 0.3041 & 0.4577 & 0.8945 &  &  &  &  &  & 
\end{tabular}}
\end{table}

\subsection{RQ2. Ablation Studies}

In this section, we evaluate the effectiveness of the main components of our model. To do this, we perform an ablation study by excluding certain main model components. The following are ablation settings that we conducted:
\begin{itemize}
  \item Without causal graphs (\textit{- Caugraph}): we exclude the causal graph generation process and instead replace it with an unweighted and fully connected graph. 
  \item Without graph convolution (\textit{- GConv}): we remove the weighted graph convolutional mechanism inside the framework.
  \item Without modified z-score (\textit{- Zscore}): we remove the modified z-score based on median absolute deviation (MAD) to standardize the anomaly scores during the inference phase.
\end{itemize}

The result of ablation studies is shown in Table~\ref{tb:ablation}. The following is the summary of the findings:
\begin{itemize}
  \item \textit{- Caugraph}: Removing the causal graph structures and instead using a fully connected graph degrades the anomaly detection performance. This demonstrates that causal graphs provide knowledge-driven structure and meaningful relationship insights between time series, thereby enhancing anomaly detection performance. Compared to \textit{- Caugraph} version, the full CGAD model achieves a 17.06 \% average improvement across all three F1 scores.
  \item \textit{- GConv}: Removing the graph convolution module decreases model performance, indicating the importance of learning graph representations through information exchange between nodes for effective anomaly detection in time series. Consequently, the complete CGAD model demonstrates an average improvement of 17.00 \% across all three F1 scores compared to the \textit{- GConv} version.
  \item \textit{- Zscore}: Removing the normalization mechanism reduces the model performance, demonstrating that the modified z-score based on MAD improves the detection process by better distinguishing the normal and abnormal patterns in multivariate time series data. The \textit{- Zscore} version exhibits a 44.37\% average decrease across all three F1 scores compared to the full CGAD model.
\end{itemize}

The ablation study results in Table~\ref{tb:ablation} demonstrate that the complete CGAD framework components are most effective for anomaly detection in multivariate time series.

\begin{figure}[!htb]
  \begin{center}
    \includegraphics[width=0.62\textwidth]{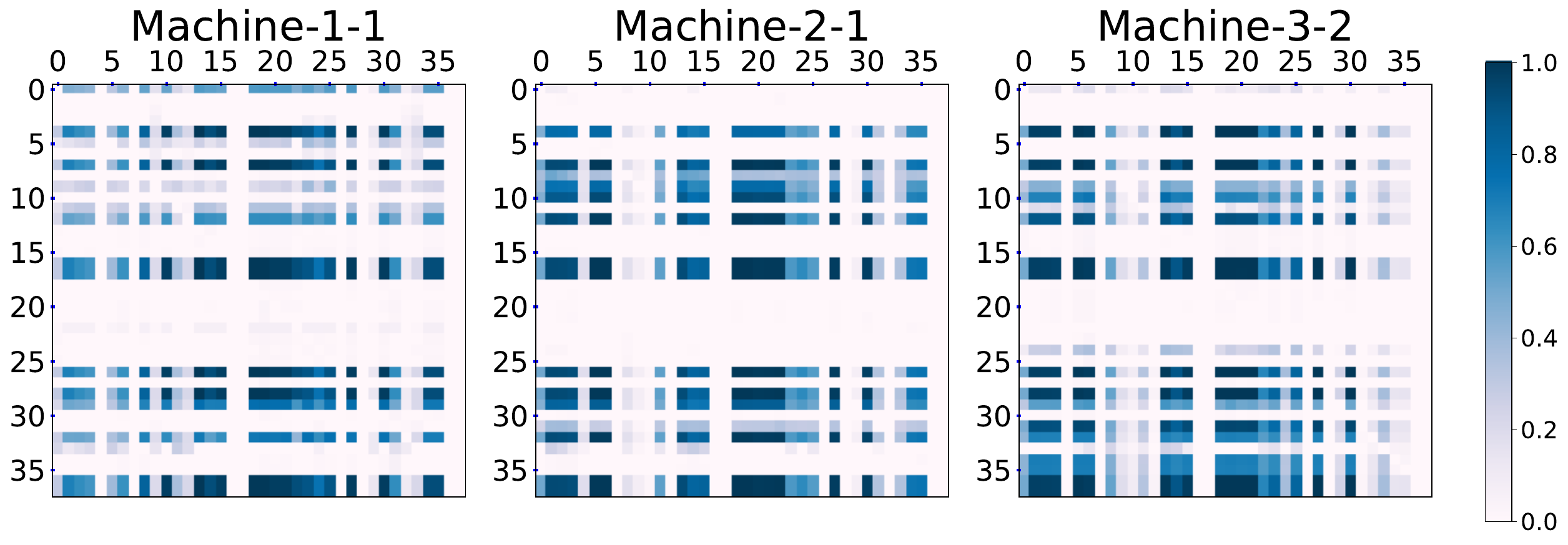}
  \end{center}
  \caption{Causal graph adjacency matrices ($38 \times 38$ nodes) of 3 subsets in the SMD dataset.}
  \label{img:adj}
\end{figure}

\begin{figure}[!htb]
  \begin{center}
    \includegraphics[width=0.82\textwidth]{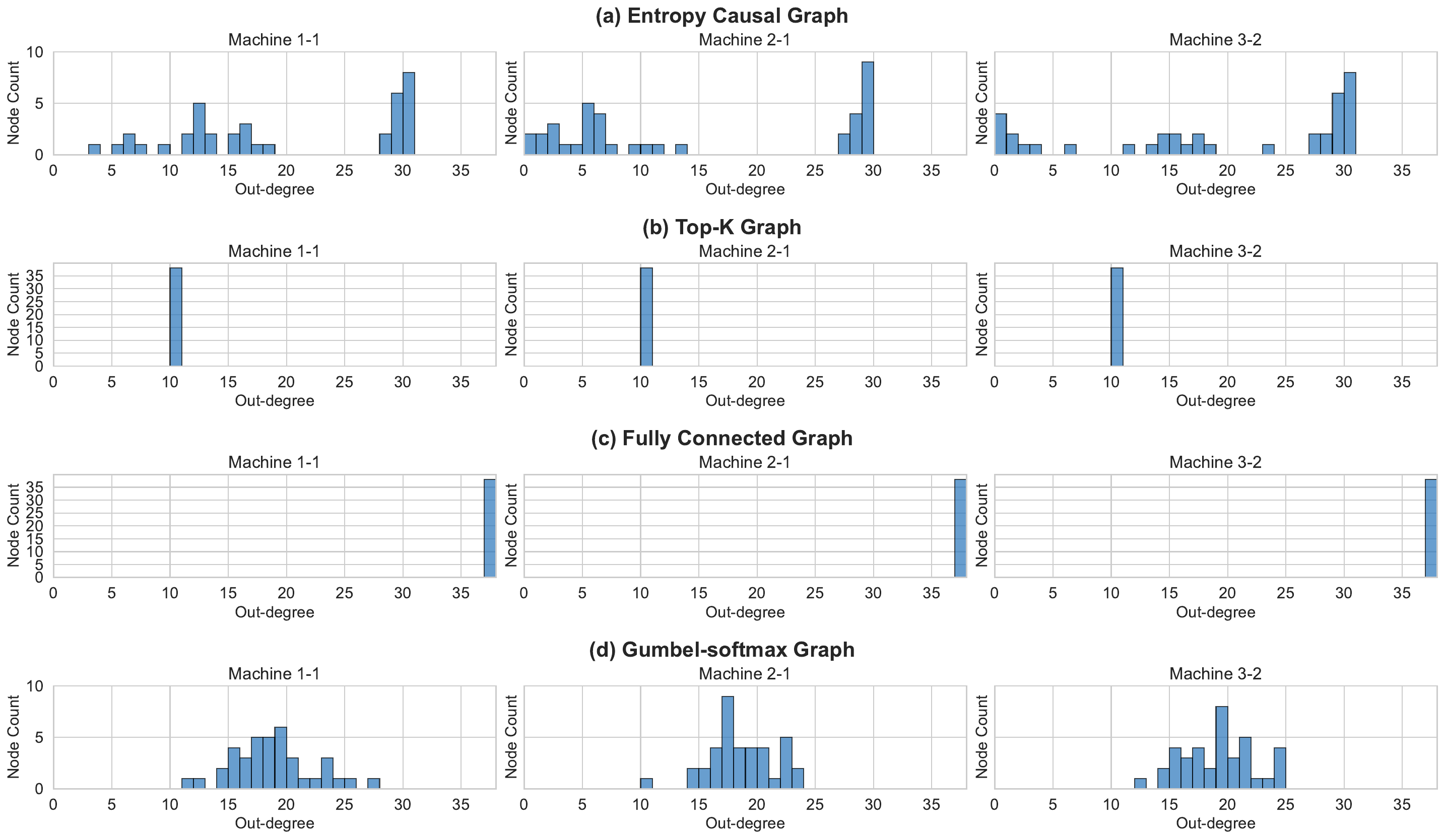}
  \end{center}
  \caption{Out-degree histogram for 3 machines in the SMD dataset.} 
  \label{img:adjdegree}
\end{figure}

\subsection{RQ3. Qualitative Analysis}
This section presents an interpretation of causal graphs in CGAD. It addresses two questions: first, whether the causal discovery method can discover relevant relations between time series data that can improve flexibility, and second, how to interpret the causal discovery process in determining the relations.

\paragraph{\textbf{Flexibility in the Causal Graph}} Fig.~\ref{img:adj} shows the SMD dataset's adjacency matrix from 3 machines, each with 38 sensors that capture stack trace data on resource utilization. The causal discovery using transfer entropy produces directed graph relations. The matrices demonstrate that causal discovery can build relevant connections only considering causality factors based on transfer entropy calculations from pairwise time series data. Specifically, the machine-1-1 subset has 727 relations from a total of 1444 possible relations (fully connected graph). 

Fig.~\ref{img:adjdegree} illustrates a comparison of the out-degree distribution from three machines within the SMD dataset across four distinct graph structures: the Entropy Causal Graph, Top-K Graph, Fully Connected Graph, and Gumbel-softmax Graph. In this context, the out-degree represents the number of edges coming out from a specific node. The histogram of out-degree resulting from the graph generation in CGAD indicates a high level of flexibility. The causal discovery uncovers only relevant relationships between time series data without restricting the graph structure to a predefined number of relations per node, as the previous methods, such as the top-$k$ or fully connected graph approach, do. In such cases, the top-$k$ approach or fully connected graph results in a histogram with only one bar, which limits the contextual information within the graph structure.
The causal graphs constructed from three machines within the SMD dataset can also reveal arbitrary patterns, as demonstrated in the out-degree histogram. This contrasts with the Gumbel-softmax graph, which consistently forms a bell-shaped or unimodal histogram. The presence of a sampling mechanism within the Gumbel-softmax, based on the Gumbel distribution, results in an out-degree histogram that consistently forms a similar bell-shaped pattern for all datasets. 
Unlike the Gumbel-softmax graph, the entropy causal graphs are more flexible and can form arbitrary shapes, as sensor connections are established only when a meaningful cause-and-effect relationship exists.

\begin{figure}[!ht]
  \begin{center}
    \includegraphics[width=0.7\textwidth]{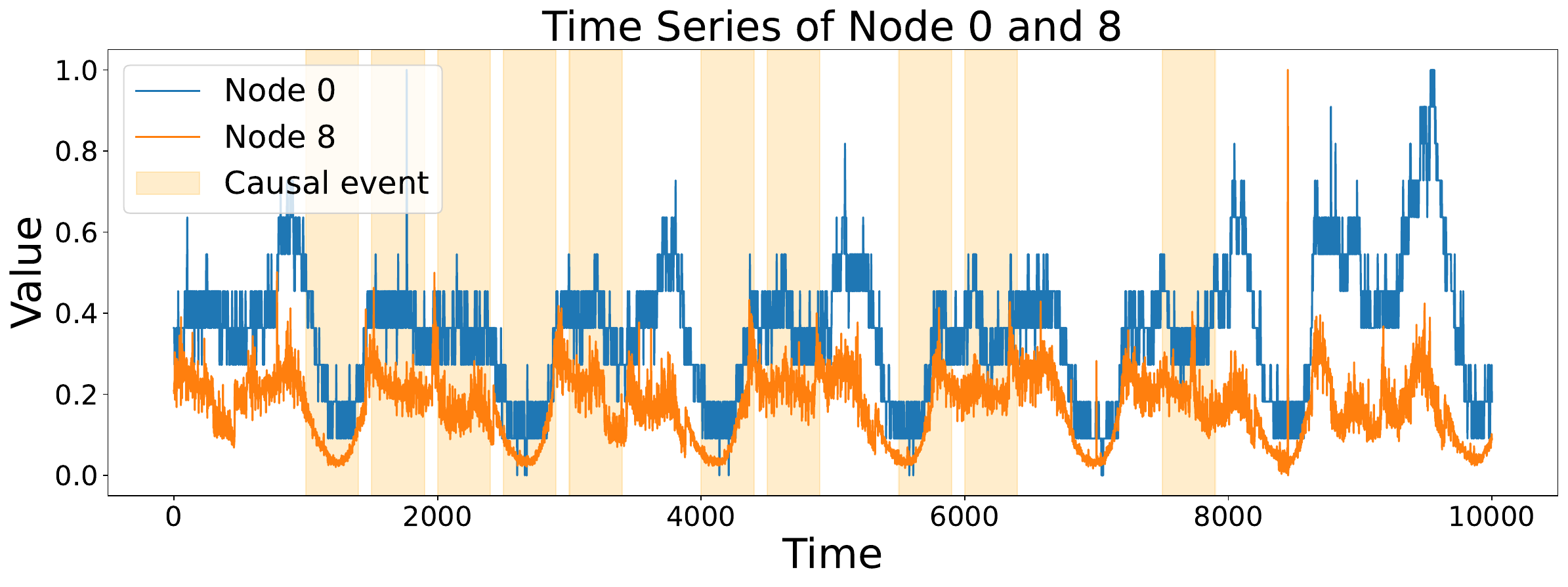}
  \end{center}
  \caption{Causal event between two nodes in Machine-1-1 subset.}
  \label{img:cau}
\end{figure}

\paragraph{\textbf{Interpretability in the Causal Graph}} The causal discovery also has better interpretability and can be visually investigated to gain a deeper understanding of underlying relationships between variables, compared to sampling techniques like the Gumbel-softmax. We use Fig.~\ref{img:cau} to answer the second question on explaining the causal relationship. The causal discovery calculation reveals that node 8 has a causal relationship with node 0. The orange blocks highlight events with high causality values. We visualize the causal events in Fig.~\ref{img:cau} by calculating the transfer entropy of two-time series data every 500-time point. Then, we highlight the top 10 causal events in every group of 500-time points to show the interpretability of the causal discovery, such as when some value decreased in node 8, node 0 also decreased in the following timeframe. Moreover, at different time steps, once the value of node 8 increases, then node 0 increases right after that event. This visually validates that node 0 has a causal relation with node 8. Overall, Table \ref{tb:ablation} together with Figures \ref{img:adjdegree} and \ref{img:cau} demonstrate that the causality enhances forecasting quality whilst providing an interpretable graph for anomaly diagnosis.

\begin{figure}[!htb]
  \begin{center}
    \includegraphics[width=0.72\textwidth]{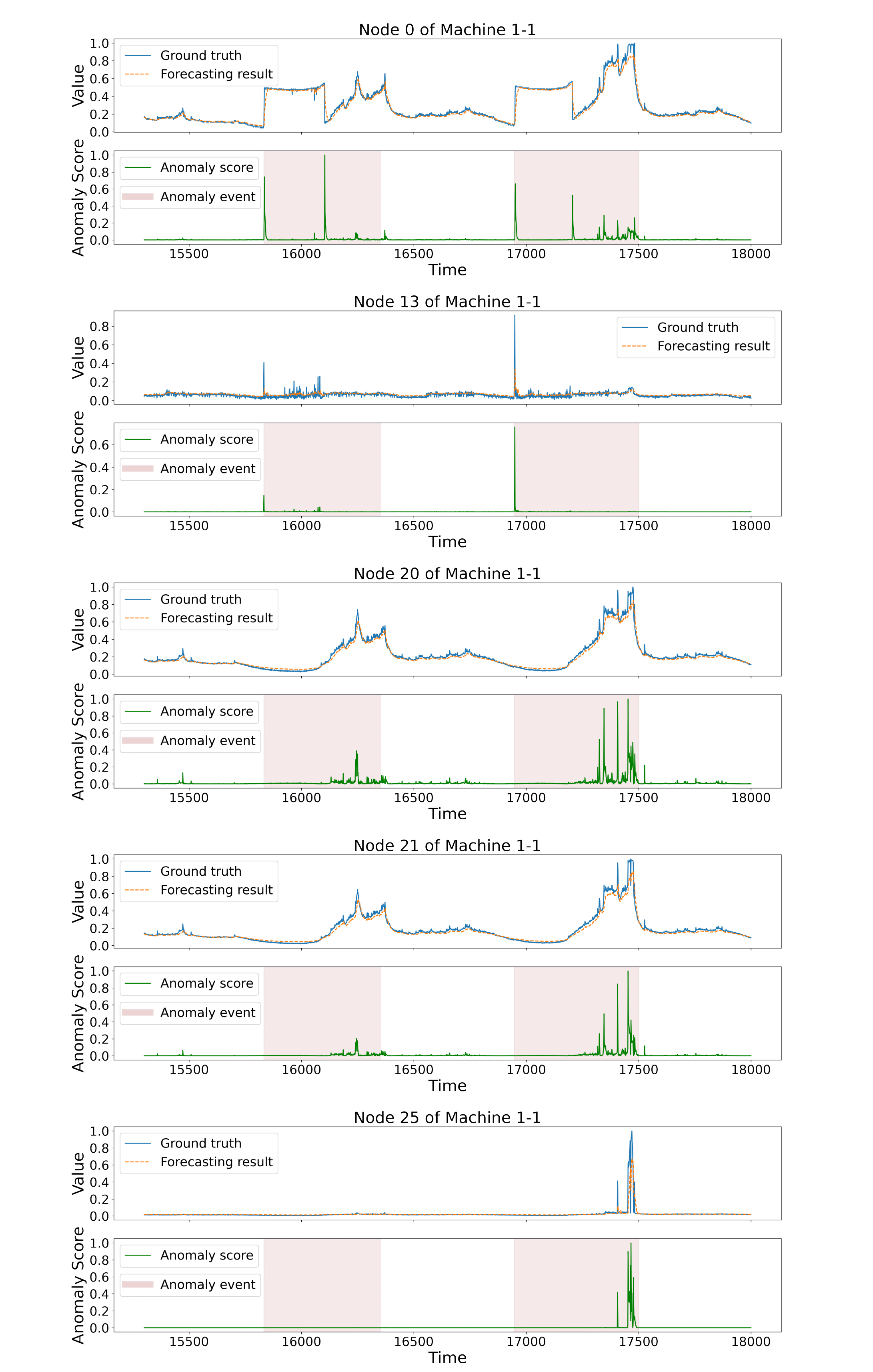
    }
  \end{center}
  \caption{Anomaly diagnosis for 5 nodes in SMD dataset. The red blocks indicate when the anomalous events occur.}
  \label{img:ano}
\end{figure}

\subsection{RQ4. Anomaly Detection Diagnosis}

This section demonstrates how the anomaly detection process in CGAD works by considering various anomaly scores across all nodes. Firstly, Fig.~\ref{img:ano} demonstrates the anomaly detection process in multivariate time series data across several nodes. For visualization purposes, we have selected a total of five nodes to be visualized from the machine 1-1 subset of the SMD dataset: Node 0 and four other nodes that exhibit a high causality factor with respect to Node 0. In each subfigure within Fig.~\ref{img:ano}, the top chart illustrates how the model forecasts time series data. The orange dashed line represents the forecasted values by the model (CGAD's forecasts), while the blue line represents the actual observed values of the system. The bottom chart in each subfigure in Fig.~\ref{img:ano} displays the corresponding anomaly scores for the time series sequences in each node. These anomaly scores are computed individually for each node by assessing the deviation between forecast predictions and ground truth labels, as shown in Equation~\ref{eq:err}. Subsequently, they are standardized using a modified z-score based on MAD, as described in Equation~\ref{eq:zscore}. These anomaly scores serve to assess the level of abnormality in specific events within the time series data. A higher anomaly score indicates that the model's prediction is inconsistent with the actual observations, suggesting a higher likelihood that an anomalous event has occurred. By examining the anomaly scores at the node level, it can be seen that the anomaly scores of certain nodes exhibit spikes during anomaly events. This information can help humans identify specific sensors or nodes that could be responsible for generating high anomaly scores.

\begin{figure}[!htb]
  \begin{center}
    \includegraphics[width=0.65\textwidth]{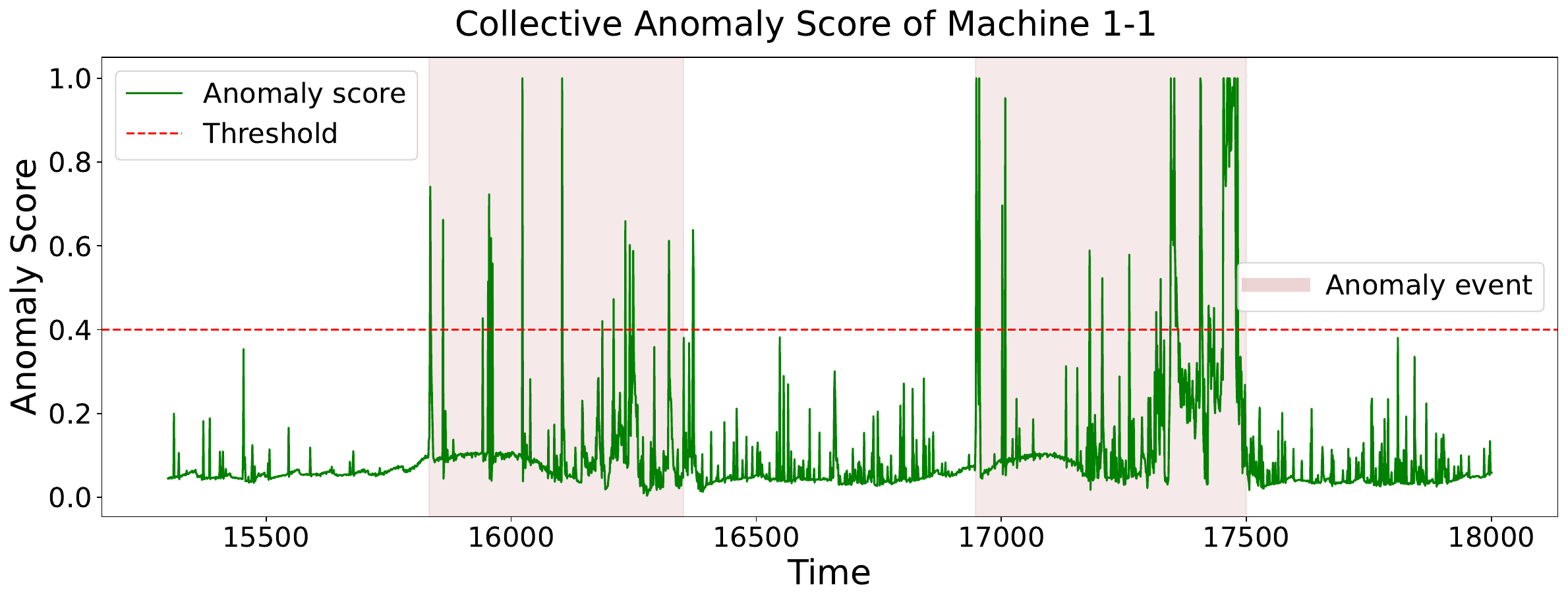}
  \end{center}
  \caption{Collective anomaly scores of all nodes in the SMD dataset. The green line represents the collective anomaly score, the dashed red line shows the anomaly detection threshold, and the red blocks indicate when the anomalous events occur.}
  \label{img:collectiveano}
\end{figure}

Fig.~\ref{img:collectiveano} represents the collective anomaly scores, which are calculated by aggregating the anomaly scores from all nodes. Namely, the anomaly score for the system at time $t$ is taken to be the largest anomaly score across all the sensors $i$ at time $t$. The anomaly alarm is triggered when the collective anomaly score exceeds the threshold, indicated by the red dashed line. We observe anomaly events occurring from steps 15900 to 16400 and 17000 to 17500. 

The performance of anomaly detection depends on the metric used. Each timestamp is considered independent of the typical point-wise score (F1), requiring a separate calculation of confusion matrix values for each timestamp. Each contiguous anomaly segment within a specific timeframe is treated as one anomaly label for the composite score ($\textup{F1}_c$). The provided example shows two anomaly labels (15900 to 16400 and 17000 to 17500). This metric helps minimize false alarms by concentrating on crucial events with a high probability of abnormality within contiguous anomaly segments at any point during the period they occur. The last metric used to calculate performance is the popular point-adjustment strategy ($\textup{F1}_{PA}$). The point-adjustment strategy labels previous and subsequent time steps anomalous if they belong to a contiguous anomaly segment while maintaining the number of time steps without any label grouping, as shown in the composite score ($\textup{F1}_c$). For example, even if some events from steps 15900 to 16400 and 17000 to 17500 have anomaly scores below the thresholds, we consider them anomalous due to their contiguous abnormal conditions.

In summary, the process of calculating anomaly scores for each node and deriving a collective anomaly score across the system forms a complete framework for diagnosing anomalies in CGAD.

\section{Conclusion}
This work proposes CGAD, an entropy causal graph for multivariate time series anomaly detection. CGAD effectively captures causal relationships among sensors or variables to enhance anomaly detection performance. The generated causal graphs based on transfer entropy provide greater flexibility without limiting relations and offer good interpretability for visually investigating causal events between time series. We use a forecasting-based strategy with GNN to model causal graphs, temporal features, and median deviation scoring for anomaly identification. Our method outperforms the state-of-the-art models, as indicated by empirical results obtained from real-world datasets in terms of the average performances of three different anomaly detection metrics, which are point-wise score (F1), composite score ($\textup{F1}_c$), and point-adjust score ($\textup{F1}_{PA}$).

\bibliographystyle{ACM-Reference-Format}
\bibliography{ref}

\end{document}